%% file: main.tex
\documentclass{article}

\usepackage{microtype}
\usepackage{graphicx}
\usepackage{subfigure}
\usepackage{booktabs} % for professional tables

\usepackage[utf8]{inputenc} % allow utf-8 input
\usepackage[T1]{fontenc}    % use 8-bit T1 fonts
\usepackage{hyperref}       % hyperlinks
\usepackage{url}            % simple URL typesetting
\usepackage{booktabs}       % professional-quality tables
\usepackage{amsfonts}       % blackboard math symbols
\usepackage{nicefrac}       % compact symbols for 1/2, etc.
\usepackage{microtype}      % microtypography
\usepackage{xcolor}         % colors

\usepackage{fixltx2e}
\usepackage{amsmath}
\usepackage{amsthm}

\usepackage{graphicx}
\usepackage{caption}
\usepackage{enumitem}
\usepackage{adjustbox}
\usepackage{pgfplots}
\usepackage{multirow}

\usepackage{ulem}
\newcommand{\tred}[1]{{#1}}

\newtheorem{remark}{Remark}

\newcommand{\bfa}{\mathbf{a}}

\newcommand{\bff}{\mathbf{f}}

\newcommand{\bfK}{\mathbf{K}}
\newcommand{\bfI}{\mathbf{I}}

\newcommand{\bfG}{\mathbf{G}}
\newcommand{\bfA}{\mathbf{A}}
\newcommand{\bfD}{\mathbf{D}}
\newcommand{\bfW}{\mathbf{W}}
\newcommand{\bfS}{\mathbf{S}}

\newcommand{\bfL}{\mathbf{L}}
\newcommand{\bfP}{\mathbf{P}}

\usepackage{hyperref}

\usepackage[accepted]{icml2023}

\usepackage{amsmath}
\usepackage{amssymb}
\usepackage{mathtools}
\usepackage{amsthm}
\usepackage{stackengine}

\usepackage[capitalize,noabbrev]{cleveref}

\theoremstyle{plain}
\newtheorem{theorem}{Theorem}[section]

\newtheorem{corollary}[theorem]{Corollary}
\theoremstyle{definition}

\usepackage[textsize=tiny]{todonotes}

\icmltitlerunning{\tred{Improving Graph Neural Networks with Learnable Propagation Operators}}

\begin{document}

\twocolumn[
%\icmltitle{$\omega$GNNs: Deep Graph Neural Networks Enhanced by Multiple Propagation Operators}
%\icmltitle{Diffuse and Sharpen: Simple and Effective Deep Graph Neural Networks}
\icmltitle{Improving Graph Neural Networks with Learnable Propagation Operators}




\begin{icmlauthorlist}
\icmlauthor{Moshe Eliasof}{bgu}
\icmlauthor{Lars Ruthotto}{em}
\icmlauthor{Eran Treister}{bgu}
\end{icmlauthorlist}

\icmlaffiliation{bgu}{Department of Computer Science, Ben-Gurion University, Beer-Sheva, Israel.}
\icmlaffiliation{em}{Department of Mathematics, Emory University, Atlanta, Georgia, USA.}

\icmlcorrespondingauthor{Moshe Eliasof}{eliasof@post.bgu.ac.il}
\icmlcorrespondingauthor{Eran Treister}{erant@cs.bgu.ac.il}

\icmlkeywords{Graph Neural Networks, oversmoothing}

\vskip 0.3in
]



\printAffiliationsAndNotice{}  % leave blank if no need to mention equal contribution

\begin{abstract}
Graph Neural Networks (GNNs) are limited in their propagation operators. In many cases, these operators often contain non-negative elements only and are shared across channels, limiting the expressiveness of GNNs. Moreover, some GNNs suffer from over-smoothing, limiting their depth. On the other hand, Convolutional Neural Networks (CNNs) can learn diverse propagation filters, and phenomena like over-smoothing are typically not apparent in CNNs.
In this paper, we bridge these gaps by incorporating trainable channel-wise weighting factors $\omega$ to learn and mix multiple smoothing and sharpening propagation operators at each layer. Our generic method is called $\omega$GNN, and is easy to implement. We study two variants: $\omega$GCN and $\omega$GAT.
For $\omega$GCN, we theoretically analyse its behaviour and the impact of $\omega$ on the obtained node features. Our experiments confirm these findings, demonstrating and explaining how both variants do not over-smooth.
Additionally, we experiment with 15 real-world datasets on node- and graph-classification tasks, where our $\omega$GCN and $\omega$GAT perform on par with state-of-the-art methods. 
\end{abstract}

\section{Introduction}
\label{sec:intro}
Graph Neural Networks (GNNs)
are useful for a wide array of fields, from computer vision and graphics \citep{monti2017geometric, wang2018dynamic, eliasof2020diffgcn} and social network analysis \citep{kipf2016semi, defferrard2016convolutional} to bio-informatics \citep{jumper2021}. Most GNNs are defined by applications of propagation and point-wise operators, where the former is often fixed and based on the graph Laplacian (e.g., GCN \citep{kipf2016semi}), or is defined by an attention mechanism \citep{velickovic2018graph, kim2021howSuperGAT, brody2022how_GATV2}.

\begin{figure}
\centering
%\begin{minipage}{0.5\linewidth}
\centering
    \includegraphics[width=1.0\linewidth]{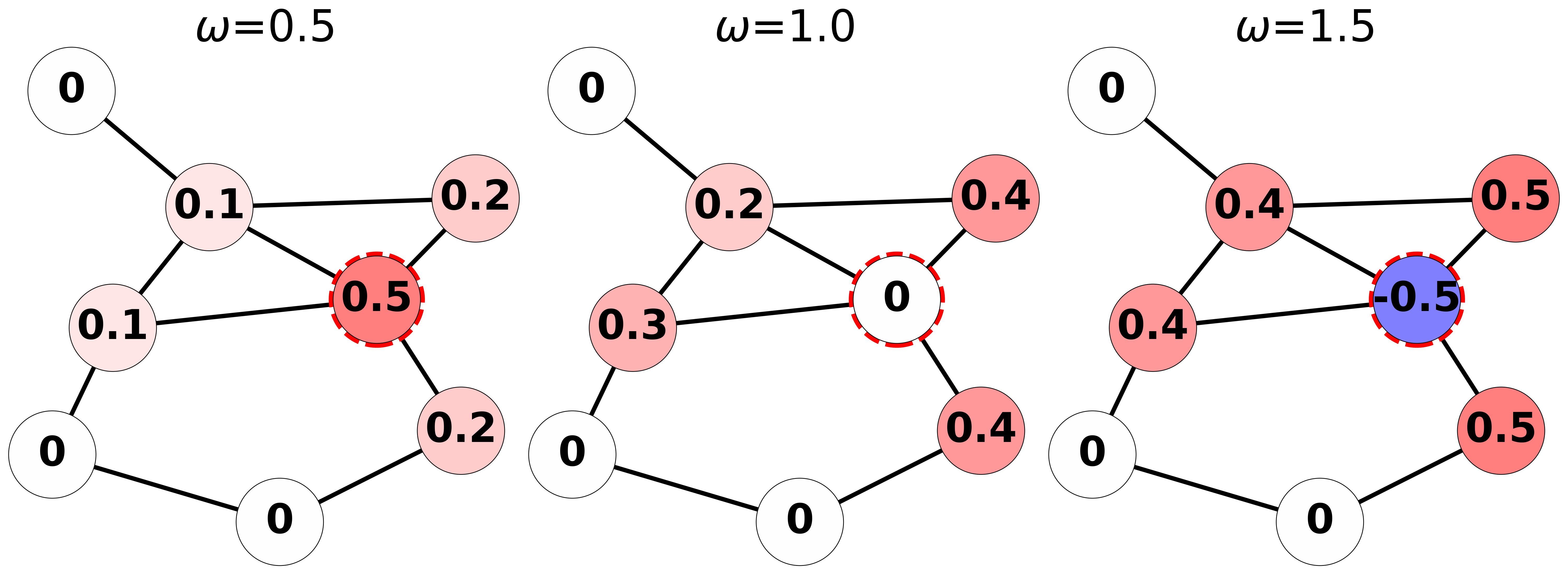}
    \caption{The impulse response of
    $\omega$GCN's propagation operator for different $\omega$ values. For $\omega=0.5,1.0$ non-negative values are obtained, while for $\omega=1.5$ we see mixed-sign values. The dashed node starts from a feature of 1 and the rest with 0.
     }
     \label{fig:omegaImpulse}
%\end{minipage}\hspace{0.5em}
\end{figure}

Most recent GNNs follow a general structure that involves two main ingredients -- the propagation operator, denoted by $\bfS^{(l)}$, and a $1\times1$ convolution denoted by $\bfK^{(l)}$, as follows
\begin{equation}
    \label{eq:general_gnn}
    \bff^{(l+1)} = \sigma(\bfS^{(l)} \bff^{(l)}\bfK^{(l)}),
\end{equation}
where $\bff^{(l)}$ denotes the feature tensor at the $l$-th layer.
%$n \times c$ 
The main limitation of the above formulation is that the propagation operators in most common architectures are constrained to be non-negative. This leads to two drawbacks. First, this limits the expressiveness of GNNs. For example, the gradient of given graph node features can not be expressed by a non-negative operator. A mixed-sign operator, as in our proposed method, can achieve this, as demonstrated in Fig. \ref{fig:omegaImpulse} and Fig. \ref{fig:gradEstimationExperiment}. Second, the utilization of  strictly non-negative propagation operators yields a smoothing process, that may lead GNNs to suffer from over-smoothing. That is, the phenomenon where node features become indistinguishable from one and other as more GNN layers are stacked -- causing severe performance degradation in deep GNNs \citep{nt2019revisiting, oono2020graph, cai2020note}.

The above drawbacks are two gaps between GNNs and
Convolutional Neural Networks (CNNs), which
can be interpreted as structured versions of GNNs (i.e., GNNs operating on a regular grid). The structured convolutions in CNNs allow to learn diverse propagation operators, and in particular, it is known that mixed-sign (high-pass) kernels like sharpening filters are useful feature extractors
in CNNs \citep{KrizhevskySutskeverHinton2012}, and such operators cannot be obtained by non-negative (smoothing) kernels only. In the context of GNNs, \citet{chien2021adaptive} have noticed that high-pass polynomial filters are learnt in their framework for heterophilic datasets, while low-pass filters are learnt for homophilic ones. A similar observation was noted in \citet{eliasof2022pathgcn} as well. In addition, the over-smoothing phenomenon is typically not evident in standard CNNs where the spatial filters are learnt, and usually adding more layers improves accuracy \citep{he2016deep}. 
% This discussion demonstrates two gaps between CNNs and GNNs that we seek to bridge in this work using simple and general GNN architectures.

A third gap between GNNs and CNNs is the ability of the latter to learn and mix multiple propagation operators. In the scope of separable convolutions, CNNs typically learn a distinct kernel per channel, known as a depth-wise convolution \citep{sandler2018mobilenetv2} -- a key element in modern CNNs \citep{tan2019efficientnet, liu2022convnet}. On the contrary, in many GNNs the propagation operator $\bfS^{(l)}$ from \eqref{eq:general_gnn} acts on all channels \citep{chen20simple, velickovic2018graph,chien2021adaptive}, and in some cases on all layers \citep{kipf2016semi,wu2019simplifying}. %in many GNNs a single propagation operator is used or learnt for the entire network \cite{kipf2016semi, wu2019simplifying}, or for each layer \cite{chen20simple, velickovic2018graph} and shared across all channels.
One exception is the multi-head GAT \citep{velickovic2018graph} where several attention heads are learnt per layer. However, this  approach  typically employs only a few heads due to the high computational cost and is still limited by learning non-negative propagation operators only. Another exception is the recent \cite{JacobiConv2022}, a linear spectral GNN that learns a different high-order (k-hop) polynomial filter per channel. In contrast, our GNN is non-linear and faithful to the form \eqref{eq:general_gnn} with a 1-hop spatial operator and $1\times 1$ convolution at each layer, similar to classical and universal CNNs. 

In this paper we propose 
% an effective modification to 
a simple and effective modification of
GNNs that closes the three gaps of GNNs discussed above, by introducing a parameter $\omega$ to control the contribution and type of the propagation operator.
We call our general approach $\omega$GNN, and 
utilize GCN \citep{kipf2016semi} and GAT \citep{velickovic2018graph} to construct two variants, $\omega$GCN and $\omega$GAT. 
We theoretically prove and empirically demonstrate that our $\omega$GNN can prevent over-smoothing. Secondly, we show that by learning $\omega$, our $\omega$GNNs can yield propagation operators with mixed signs, ranging from smoothing to sharpening operators (see Fig. \ref{fig:omegaImpulse} for an illustration). This approach enhances the expressiveness of the network, as demonstrated in Fig. \ref{fig:gradEstimationExperiment}, in contrast to many other GNNs that employ non-negative (smoothing) propagation operators only.
Lastly, we propose and demonstrate that by learning different $\omega$ per layer and channel, similarly to a depth-wise convolution in CNNs, our $\omega$GNNs obtain competitive accuracy.

Our contributions are summarized as follows:\vspace{-3pt} 
\begin{itemize}[leftmargin=2em]\setlength\itemsep{0em}
    \item We propose $\omega$GNN, an effective and computationally light modification to GNNs of a common and generic structure, that directly avoids over-smoothing and enhances the expressiveness of GNNs. Our method is demonstrated by $\omega$GCN and $\omega$GAT.
    \item A theoretical analysis and experimental validation of the behaviour of $\omega$GNN are provided to expose its improved expressiveness compared to standard propagation operators in GNNs. 
    \item We propose to learn multiple propagation operators by learning $\omega$ \emph{per layer and per channel} and mixing them using a $1\times1$ convolution followed by a non-linear activation) to enhance the expressiveness of GNNs.
    \item Our experiments with 15 real-world datasets on numerous applications and settings, from semi- and fully-supervised node classification to graph classification show that our $\omega$GCN and $\omega$GAT read performance on par with current state-of-the-art methods.
\end{itemize}

%\begin{minipage}{0.48\linewidth}
\begin{figure}
    \centering
    \begin{subfigure}{}
    \footnotesize
    \stackon[5pt]{
    \includegraphics[width=0.165\linewidth, height=0.165\linewidth]
    {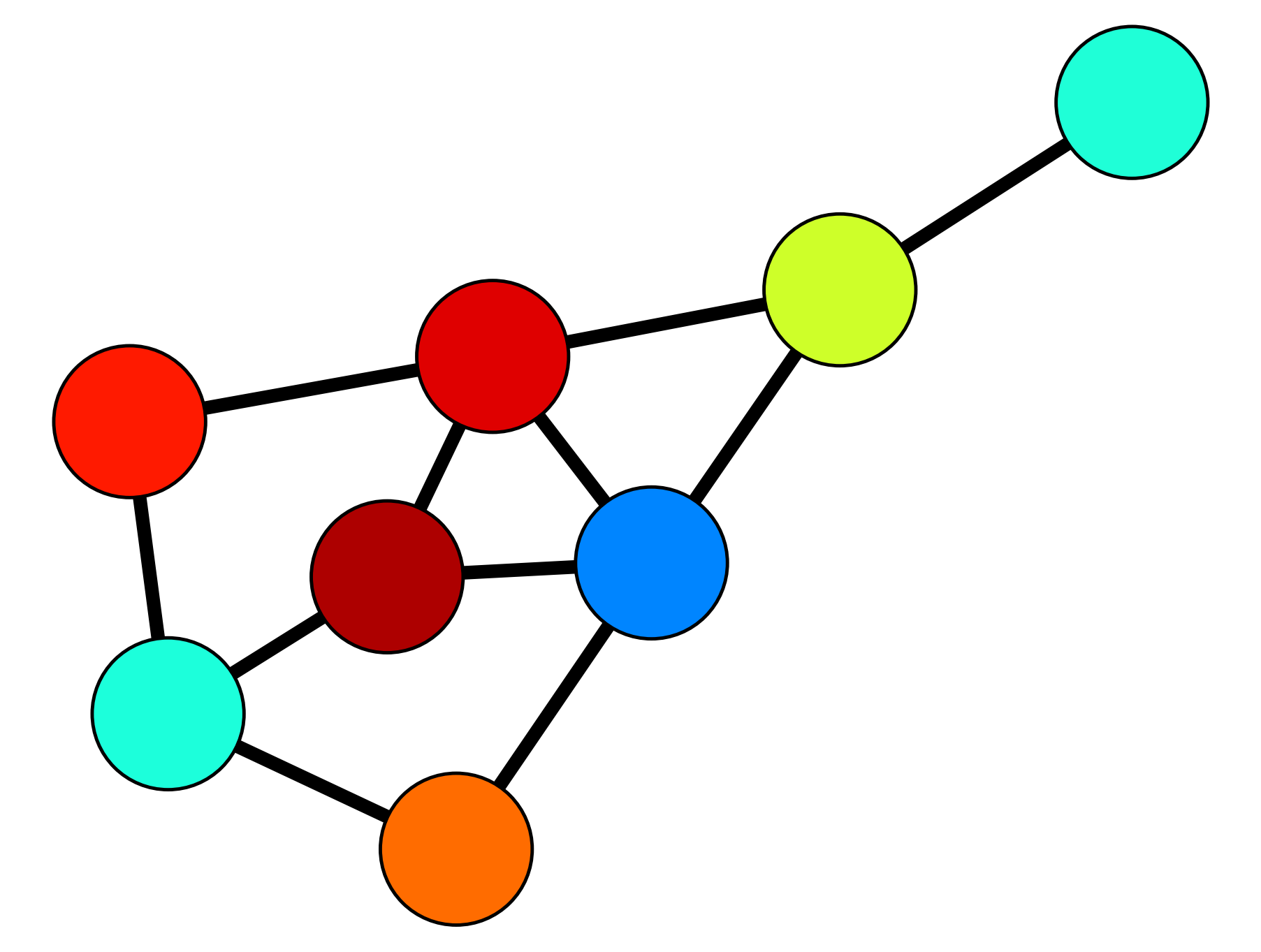}}{Input}
    %\caption{Input graph.}
    \end{subfigure}
    %\\
    \begin{subfigure}{}
    \footnotesize
    \stackon[5pt]{
    \includegraphics[width=0.165\linewidth, height=0.165\linewidth]{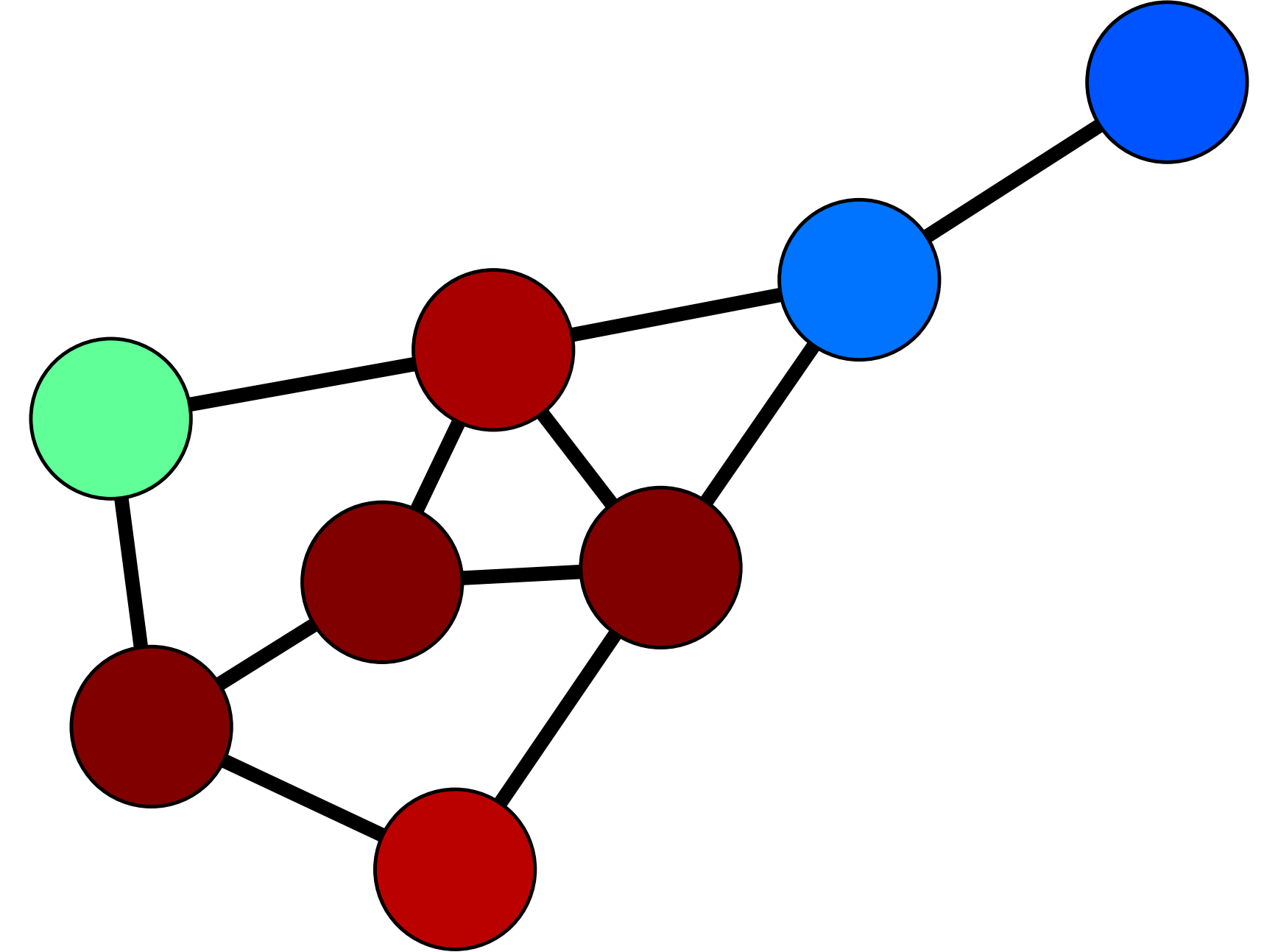}}{Gradient}
        %\caption{Node gradient.}
    \end{subfigure}
    \begin{subfigure}{}
    \footnotesize
    \stackon[5pt]{\includegraphics[width=0.165\linewidth, height=0.165\linewidth]{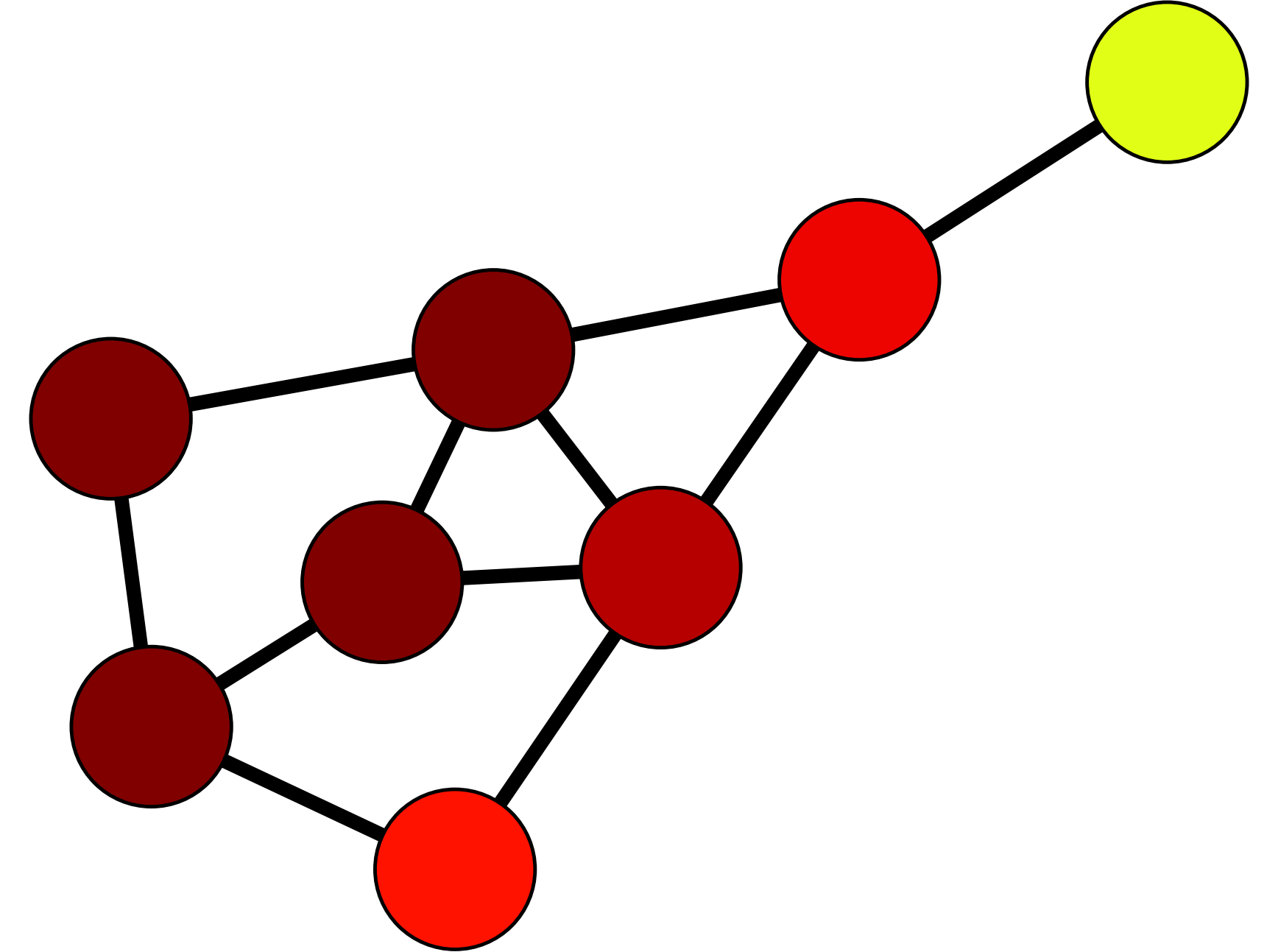}}{GCN}
        %\caption{GCN estimation.}
    \end{subfigure}
        \begin{subfigure}{}
    \footnotesize
    \stackon[5pt]{
    \includegraphics[width=0.165\linewidth, height=0.165\linewidth]{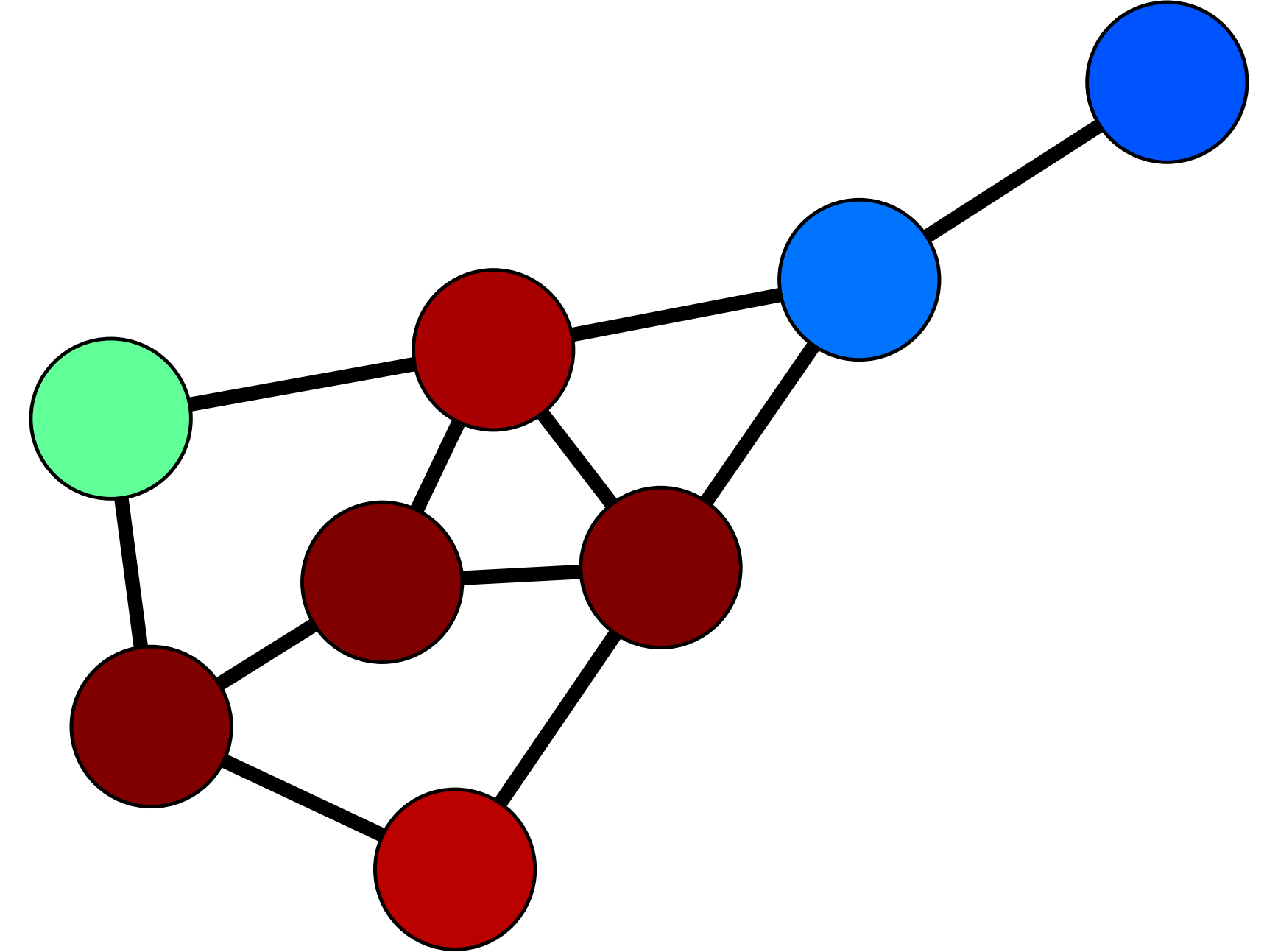}}{$\omega$GCN}
    %\caption{$\omega$GCN estimation.}
    \end{subfigure}
    \begin{subfigure}{}
    \includegraphics[width=0.125\linewidth, height=0.2\linewidth, trim={10cm 0cm 0cm 0},clip]{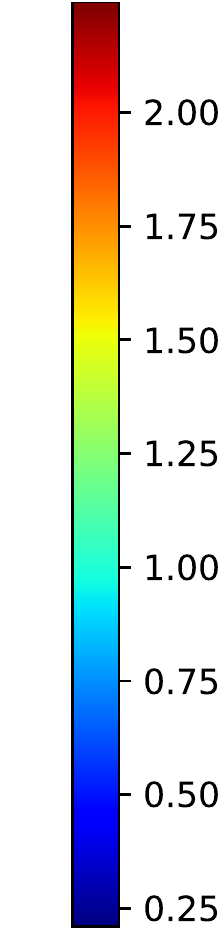}
    \end{subfigure}
    \caption{The expressiveness of $\omega$GNNs. Our $\omega$GCN can estimate the gradient of the node features while GCN cannot.}
\label{fig:gradEstimationExperiment}
%\end{minipage}
\end{figure}

\iffalse
\begin{figure}[h]
    \centering
    \begin{subfigure}{0.24\linewidth}
    \includegraphics[width=1\linewidth]
    {exp_figs/gtSignal.eps}
    \caption{Input graph and node features (coloured).}
    \end{subfigure}
    %\\
    \begin{subfigure}{0.24\linewidth}
    \includegraphics[width=1\linewidth]{exp_figs/gtGrad.eps}
        \caption{The gradient of the nodes features.}
    \end{subfigure}
    \begin{subfigure}{0.24\linewidth}
    \includegraphics[width=1\linewidth]{exp_figs/gcnGradPred.eps}
        \caption{Estimated node features by GCN.}
    \end{subfigure}
        \begin{subfigure}{0.24\linewidth}
    \includegraphics[width=1\linewidth]{exp_figs/wgcnGradPred.eps}
        \caption{Estimated node features by $\omega$GCN.}
    \end{subfigure}
    \caption{The expressiveness of $\omega$GNNs. Our $\omega$GCN can express the gradient function of the node features while GCN cannot.}
    \label{fig:gradEstimationExperiment}
\end{figure}
\fi

%\paragraph{Graph Pooling.}
%\cite{ying2018hierarchical,gao2019graph, knyazev2019understanding, ranjan2019asap, diehl2019edge, Khasahmadi2020Memory-Based}

\section{Method}
\label{sec:wgnns}
We start by providing the notations that will be used throughout this paper, and displaying our general $\omega$GNN in Sec. \ref{sec:wgnns_details}. Then we consider two popular GNNs that adhere to the structure presented in  \eqref{eq:general_gnn}, namely GCN and GAT. We formulate and analyse the behaviour of their two counterparts $\omega$GCN and $\omega$GAT in Sec. \ref{sec:omegaGCN} and \ref{sec:omegaGAT}, respectively. %GCN and GAT are two architectures which have been shown to over-smooth in the literature \cite{li2018deeper, wu2019simplifying, Zhao2020PairNorm:, measuringoversmoothing},

\paragraph{Notations.} Assume we are given an undirected graph defined by the set $\cal G=({\cal V},{\cal E})$  where $\cal V$ is a set of $n$ vertices and $\cal E$ is a set of $m$ edges. Let us denote by ${\bf f}_i\in\mathbb{R}^c$ the feature vector of the $i$-th node of $\cal G$ with $c$ channels. Also, we denote the adjacency matrix $\bfA$, where $\bfA_{ij} = 1$ if there exists an edge $(i,j) \in {\cal E}$ and 0 otherwise. We also define the diagonal degree matrix $\bfD$ where $\bfD_{ii}$ is the degree of the $i$-th node. The graph Laplacian is given by $\bfL=\bfD-\bfA$.
Let us also denote the adjacency and degree matrices with added self-loops by $\tilde \bfA$ and $\tilde \bfD$, respectively. 
Lastly, we denote the symmetrically normalized graph Laplacian by $\tilde \bfL^{sym} = \tilde{\bfD}^{-\frac{1}{2}}\tilde{\bfL}\tilde{\bfD}^{-\frac{1}{2}}$ where $\tilde{\bfL} = \tilde{\bfD} - \tilde{\bfA}$.

\subsection{$\omega$GNNs}
\label{sec:wgnns_details}

The goal of $\omega$GNNs is to utilize learnable mixed-sign propagation operators that control smoothing and sharpening to enrich GNNs expressiveness. Below, we describe how the learnt $\omega$ influences the obtained operator and how to learn and mix multiple operators for enhanced expressiveness.

\paragraph{Learning propagation weight $\omega$.} To address the expressiveness and over-smoothing issues, we suggest a general form given an arbitrary non-negative and normalized (e.g., such that its row sums equal to 1) propagation operator $\bfS^{(l)}$. Our general $\omega$GNN is then given by
\begin{equation}
	\label{eq:omegaGNN}
	\bff^{(l+1)} = \sigma \left(\left(\bfI - {\omega}^{(l)} \left(\bfI - \bfS^{(l)}\right)\right)\bff^{(l)}\bfK^{(l)}\right),
\end{equation}
where $\omega^{(l)}$ is a scalar that is learnt per layer, and in the next paragraph we offer a more elaborated version with a parameter $\omega$ per layer and channel. The introduction of $\omega^{(l)}$ allows our $\omega$GNN layer to behave in a three-fold manner.
When $\omega^{(l)} \leq 1$, a smoothing process is obtained \footnote{The use of the value 1 in this discussion corresponds to a non-negative operator $\bfS^{(l)}$ with zeros on its diagonal, normalized to have row sums of 1. Other normalizations may yield other constants. Also, if $0<\bfS^{(l)}_{ii}<1$, then setting $\omega^{(l)}>\frac{1}{1-\bfS^{(l)}_{ii}}$ flips the sign of the $i$-th diagonal entry. }. Note, that for $\omega^{(l)}=1$,  \eqref{eq:omegaGNN} reduces to the standard GNN dynamics from  \eqref{eq:general_gnn}.
%, and if $\bfS^{(l)} = \tilde{\bfP}$ it reduces to GCN \cite{kipf2016semi} as in  \eqref{eq:gcn}.
In case ${\omega}^{(l)}=0$,  \eqref{eq:omegaGNN} reduces to a $1 \times 1$ convolution followed by a non-linear activation function, and does not propagate neighbouring node features. %In this case $\bfS=\bfI$ and only a $1 \times 1$ convolution is performed, and all the eigenvalues of $\bfI$ are equal, that is, there is no leading eigenvector. Hence, the over-smoothing phenomenon does not occur.
On the other hand, if $\omega^{(l)} > 1$, we obtain an operator with negative signs on the diagonal but positive on the off-diagonal entries, inducing a sharpening operator.
An example of various $\omega^{(l)}$ values and their impulse response is given in Fig. \ref{fig:omegaImpulse}. 
Thus, a learnable $\omega^{(l)}$ allows to learn a new family of operators, namely sharpening operators, that are not achieved by methods like GCN and GAT. To demonstrate the importance of sharpening operators, we consider a synthetic task of node gradient feature regression, given a graph and input node features(see Appendix \ref{appendix:synthetic_task} for more details). As depicted in Fig. \ref{fig:gradEstimationExperiment}, using a non-negative operator as in GCN cannot accurately express the gradient operator output, while our $\omega$GCN estimates the gradient output with a machine precision accuracy. Also, the benefit of employing both smoothing and sharpening operators is reflected in the obtained accuracy of our method on real-world datasets in Sec. \ref{sec:experiments}.

\paragraph{Multiple propagation operators.}
To learn multiple propagation operators, we extend  \eqref{eq:omegaGNN} from a channels-shared weight to channel-wise weights by learning a vector $\vec{\omega}^{(l)}\in c$ as follows
\begin{equation}
	\label{eq:omegaGNN_perChannel}
	\bff^{(l+1)} = \sigma\left(\left(\bfI - {\boldsymbol{\Omega}_{\vec{\omega}^{(l)}}} \left(\bfI - \bfS^{(l)}\right)\right)\bff^{(l)}\bfK^{(l)}\right),
\end{equation}
%\sout{where $\boldsymbol{\Omega}_{\vec{\omega}^{(l)}} = Diag(\vec{\omega}^{(l)})$ is a diagonal matrix with $\vec{\omega}^{(l)}$ on its diagonal.}
where $\boldsymbol{\Omega}_{\vec{\omega}^{(l)}}$ is an operator that scales each channel $j$ with a different $\omega_j^{(l)}$.
As discussed in Sec. \ref{sec:intro}, this procedure yields a propagation operator per-channel, which is similar to depth-wise convolutions in CNNs \citep{howard2017mobilenets, sandler2018mobilenetv2}. Thus, the extension to a vector $\vec{\omega}^{(l)}$ helps to further bridge the gap between GNNs and CNNs.

We note that using this approach, our $\omega$GNN is suitable to many existing GNNs, and in particular to those which act as a separable convolution, as described in Eq. \eqref{eq:general_gnn}. In what follows, we present and analyse two variants based on GCN and GAT, called $\omega$GCN and $\omega$GAT, respectively.

\subsection{$\omega$GCN}
\label{sec:omegaGCN}
\iffalse
First, we observe that the propagation operator in GCN \citep{kipf2016semi} can be written as
\begin{equation}
\tilde{\bfP} = \tilde\bfD^{-\frac{1}{2}} \bfA\tilde\bfD^{-\frac{1}{2}} =  \bfI - \left(\bfI - \tilde\bfD^{-\frac{1}{2}} \bfA\tilde\bfD^{-\frac{1}{2}}\right). %=  \bfI -  \tilde\bfD^{-\frac{1}{2}} \bfL\tilde\bfD^{-\frac{1}{2}}.
\end{equation}
Therefore, a GCN layer reads
\begin{equation}
    \label{eq:GCNasJacobi}
    \bff^{(l+1)} =  \sigma\left(\left(\bfI - \left(\bfI - \tilde\bfD^{-\frac{1}{2}} \bfA\tilde\bfD^{-\frac{1}{2}}\right)\right)\bff^{(l)}\bfK^{(l)}\right), % = \sigma((\bfI -  \tilde\bfD^{-\frac{1}{2}} \bfL\tilde\bfD^{-\frac{1}{2}})\bff^{(l)}\bfK^{(l)}),
\end{equation}
which is a smoothing process governed by the GCN propagation operator\footnote{See Appendix \ref{appendix:proofs} for a derivation.} $\tilde{\bfP}$. Thus, repeated applications of \eqref{eq:GCNasJacobi} will lead to the over-smoothing phenomenon, as shown in \cite{wu2019simplifying,wang2019improving}. 

\fi

GCNs are a class of GNNs that employ a pre-determined propagation operator $\tilde\bfP=
\tilde{\bfD}^{-\frac{1}{2}}\tilde{\bfA}\tilde{\bfD}^{-\frac{1}{2}}$, that stems from the graph Laplacian. For instance, GCN \cite{kipf2016semi} is given by:
\begin{equation}
    \label{eq:gcn}
    \bff^{(l+1)} = \sigma(\tilde{\bfP} \bff^{(l)}\bfK^{(l)}),
\end{equation}
that is, by setting $\bfS^{(l)} = \tilde{\bfP}$ in  \eqref{eq:general_gnn}.

Other methods like SGC \citep{wu2019simplifying}, GCNII \citep{chen20simple} and EGNN \citep{zhou2021dirichlet} also rely on $\tilde{\bfP}$ as a propagation operator. 

The operator $\tilde{\bfP}$ is a fixed non-negative smoothing operator, hence, repeated applications of \eqref{eq:gcn} lead to the over-smoothing phenomenon, where the feature maps converge to a single eigenvector as shown by \cite{wu2019simplifying,wang2019improving}. Moreover, $\tilde{\bfP}$ is pre-determined, and solely depends on the graph connectivity, disregarding the node features, which may harm performance.

By baking our proposed $\omega$GNN with a  learnable weight, denoted by ${\omega}^{(l)} \in \mathbb{R}$ into GCN we obtain the following propagation scheme, named $\omega$GCN:
\begin{equation}
    \label{eq:GCNasJacobiLearn}
    \bff^{(l+1)} = 
    \sigma\left(\left(\bfI - \omega^{(l)}\left(\bfI - \tilde\bfP\right)\right)\bff^{(l)}\bfK^{(l)}\right).
\end{equation}
We now present theoretical analyses of our $\omega$GCN and reason about its \emph{non} over-smoothing property.
 We first define the node features Dirichlet 
energy at the $l$-th layer, as in \cite{zhou2021dirichlet}:
\begin{equation}
    \label{eq:dirichletEnergy}
    E(\bff^{(l)}) = \sum_{i \in \mathcal{V}} \sum_{j \in \mathcal{N}_i} \frac{1}{2} \textstyle{\left\|\frac{\bff^{(l)}_i}{\sqrt{(1+d_i)}} - \frac{\bff^{(l)}_j}{\sqrt{(1+d_j)}} \right\|_2^2}. %= \frac{1}{2}\|\bfG\tilde\bfD^{-\frac{1}{2}}\bff^{(l)}\|_2^2,
\end{equation}
%where $\bfG$ is the graph gradient operator, also known as the incidence matrix, that for each edge subtracts the features of the two connected nodes, i.e., $\bfG \bff^{(l)}_{(i,j)} = \bff^{(l)}_i - \bff^{(l)}_j$ for $(i,j)\in\mathcal{E}$.
Fig. \ref{fig:energyFig1} demonstrates how the Dirichlet energy $E(\bff^{(l)})$ decays to zero when $\omega$ is a constant, and to a fixed positive value %\sout{and how $E(\bff^{(l)})$ does not decay} 
when $\omega$ is learnt.
Next, we provide a theorem that characterizes the behaviour of $\omega$ and how it prevents over-smoothing. To this end we denote the propagation operator of $\omega$GCN from  \eqref{eq:GCNasJacobiLearn} by
\begin{eqnarray}\label{eq:P_omega}
\tilde\bfP_\omega &=& \bfI - \omega\left(\bfI - \tilde\bfP\right) \\&=& \bfI - \omega\left(\bfI - \tilde\bfD^{-\frac{1}{2}} \tilde\bfA\tilde\bfD^{-\frac{1}{2}}\right) \\&=& \bfI - \omega \tilde\bfD^{-\frac{1}{2}} \bfL\tilde\bfD^{-\frac{1}{2}},
\end{eqnarray}
where the latter equality is shown in Appendix \ref{appendix:proofs}. In essence, we show that repeatedly applying the operator $\tilde\bfP$ is equivalent to applying gradient descent steps for minimizing  \eqref{eq:dirichletEnergy} with a learning rate $\omega$. We build on the observation that smoothing is beneficial \citep{gasteiger_diffusion_2019, chamberlain2021grand} and assume that for each dataset there exists a plausible energy value at the last layer that satisfies $0<E(\bff^{(L)}) < E(\bff^{(0)})$. We note that if the graph is strongly connected, then if $E(\bff^{(L)})=0$ it means that all the feature maps are constant at the output of the network. It is reasonable to expect that such constant feature maps will not yield good performance, at least in tasks like node classification, for example. 

\tred{Now, assuming that we wish to have some energy $E^*$ at the last layer, }we show that if we learn a single $\omega^{(l)} = \omega>0$, shared across all layers, then taking $L$ to infinity will lead the learned $\omega$ to zero, \tred{scaling as $1/L$}. Thus, our $\omega$GCN will not over-smooth, as the energy at the last layer $E(\bff^{(L)})$  can reach to $E_{L}$. Later, in Corollary \ref{cor:sumOmega}, we generalize this result for a per-layer $\omega^{(l)}$, and empirically validate both results in Sec. \ref{sec:ablation} and Fig. \ref{fig:sumOmega}. The proofs for the Theorem and Corollary below are given in Appendix \ref{appendix:proofs}.
\input{images/FigEnergy}

\begin{theorem}
\label{theorem:single_omega}
Consider $L$ applications of \eqref{eq:P_omega}, i.e., $\bff^{(L)} = (\tilde{\bfP}_{\omega})^{L} \bff^{(0)}$ with a shared parameter $\omega^{(l)}=\omega$ that is used in all the layers. \tred{Also, assume that there is some desired Dirichlet energy $E(\bff^{(L)})=E^*$ of the final feature map that satisfies $0<E^*<E(\bff^{(0)})$. Then, at the limit, as more layers are added and $L$ grows, the value of the learnt $\omega$ converges to $T/L$ for some constant $T$%product $L\omega$ using the learned $\omega$ converges to constant independent of $L$
, up to first-order accuracy.
}
\end{theorem}

% \begin{theorem}
% \label{theorem:single_omega}
% Let $E^* \in (0, E(\bff^{(0)}))$ be  given and $L$ be the number of layers.
% Then, there exists $T$ such that with  shared parameter $\omega^{(l)}=\frac{T}{L}$, we have that $E(f^{(L)}) = E^* + \mathcal{O}(1/L)$ where 
% $\bff^{(L)} = (\tilde{\bfP}_{\omega})^{L} \bff^{(0)}$ .

% % the Dirichlet energy of  satisfies 
% % $E^* = $
% % Consider L applications of \eqref{eq:P_omega}, i.e., $\bff^{(L)} = (\tilde{\bfP}_{\omega})^{L} \bff^{(0)}$ with a shared parameter $\omega^{(l)}=\omega$ that is used in all the layers. \tred{Also, assume that there is some desired Dirichlet energy $E^*$ of the final feature map that satisfies $0<E^*<E(\bff^{(0)})$. Then, at the limit, as more layers are added and $L$ grows, the value of the product $L\omega$ using the learned $\omega$ converges to constant independent of $L$, up to first-order accuracy.
% % }
% \end{theorem}

\begin{corollary}
\label{cor:sumOmega}
Allowing a variable $\omega^{(l)} > 0$ at each layer in Theorem \ref{theorem:single_omega}, yields that $\sum_{l=0}^{L-1} \omega^{(l)}$ converges to a constant independent of $L$ up to first order accuracy.
\end{corollary}

Next, we dwell on the second mechanism in which $\omega$GCN prevents over-smoothing.
We analyse the eigenvectors of $\tilde{\bfP}_\omega$, showing that different choices of $\omega$ yield different leading eigenvectors that alter the behaviour of the propagation operator (i.e. smoothing and sharpening processes). This result is useful because
changing the leading eigenvector prevents the gravitation towards a specific eigenvector, which causes the over-smoothing to occur \citep{wu2019simplifying, oono2020graph}.
\begin{theorem}
\label{theorem:omega_smoothing}
Assume that the graph is connected. Then, there exists some $\omega_0\geq1$ where for all $0<\omega<\omega_0$, the operator $\tilde\bfP_\omega$ in  \eqref{eq:P_omega} is smoothing and the leading eigenvector is  $\tilde\bfD^{\frac{1}{2}}\mathbf{1}$. For $\omega > \omega_0$ or $\omega < 0$, the leading eigenvector changes.
\end{theorem}

The proof for the theorem is given in Appendix  \ref{appendix:proofs}.
%\begin{remark}[The non-negativity of $\tilde\bfP_\omega$]
%By definition, for $0 < \omega \leq 1$ all the spatial weights of     $\tilde\bfP_\omega$ defined in Eq. \eqref{eq:P_omega} are non-negative, and it is that the operator is smoothing as it is a low-pass filter. For $\omega > 1$ or $\omega < 0$, by definition we have an operator with mixed signs. 
%\end{remark}

%As discussed earlier, the proposed formulation in \eqref{eq:GCNasJacobiLearn} is more expressive than the one in  \eqref{eq:gcn} and in particular can lead to a plethora of propagation operators, from smoothing to sharpening and edge-detecting kernels. 
\textbf{$\omega$GCN with multiple propagation operators.} To further increase the expressiveness of our $\omega$GCN we extend ${\omega}^{(l)} \in \mathbb{R}$ to $\vec{\omega}^{(l)} \in \mathbb{R}^{c}$ and learn a propagation operator per channel, at each layer. %Such multiple spatial kernels are learnt and known to be useful (see discussions in \cite{howard2017mobilenets, sandler2018mobilenetv2}).
To this end, we modify \eqref{eq:GCNasJacobiLearn} to the following formulation
\begin{equation}
    \label{eq:omegaLearnMultiple}
        \bff^{(l+1)} = %\sigma\left(\left(\bfI - \Omega_{\vec{\omega}^{(l)}}\left(\tilde\bfD^{-\frac{1}{2}} \bfL\tilde\bfD^{-\frac{1}{2}}\right)\right)\bff^{(l)}\bfK^{(l)}\right).
        \sigma\left(\left(\bfI - \boldsymbol{\Omega}_{\vec{\omega}^{(l)}}\left(\bfI-\tilde{\bfP}\right)\right)\bff^{(l)}\bfK^{(l)}\right).
\end{equation}
%where $\Omega^{(l)}$ is a block diagonal matrix that scales each channel with a different $\omega$ and is parameterized by a learnable vector of $c$ scalars, where $c$ is the number of channels. 
As we show in Sec. \ref{sec:ablation}, learning a propagation operator per channel is beneficial to improve accuracy. %That is, we effectively learn a different spatial operator \emph{per-channel}, yielding a more expressive learnt set of filters to improve performance.

\subsection{$\omega$GAT}
\label{sec:omegaGAT}

%Learn a non-negative weight to every edge $(i, j) \in \mathcal{E}$, which  acts as a propagation operator and replaces $\tilde{\bfP}$.
The seminal GAT \citep{velickovic2018graph} learns a non-negative edge-weight as follows
\begin{equation}
\small
    \label{eq:attentionCoefficients}
    \alpha_{ij}^{(l)} = \frac{\exp \big({\rm LeakyReLU} \big(\bfa^{(l)^{\top}} [\bfW^{(l)} \bff_i^{(l)} || \bfW^{(l)} \bff_j^{(l)} ]  \big) \big)}{\sum_{p \in \mathcal{N}_i} \exp \big({\rm LeakyReLU} \big(\bfa^{(l)^{\top}} [\bfW^{(l)} \bff_i^{(l)} || \bfW^{(l)} \bff_p^{(l)} ]  \big) \big)},
\end{equation}
where $\bfa^{(l)} \in \mathbb{R}^{2c}$ and $\bfW^{(l)} \in \mathbb{R}^{c \times c}$ are trainable parameters and $||$ denotes channel-wise concatenation. %Then, the $l$-th GAT layer can be written as:
%\begin{equation}
%    \label{eq:gat}
%    \bff^{(l+1)}_i =  \sigma ( \underset{j \in \mathcal{N}_i}{\sum} \alpha_{ij}^{(l)} \bfW^{(l)} \bff^{(l)}_i ) = \sigma(\bfA^{(l)}\bff_i^{(l)}\bfW^{(l)}).
%\end{equation}
%\begin{equation}
%    \label{eq:gat}
%    \bff^{(l+1)} = \sigma(\bfA^{(l)}\bff^{(l)}\bfW^{(l)}).
%\end{equation}
Here, GAT is obtained by defining the propagation operator $\bfS^{(l)}$ in  \eqref{eq:general_gnn} as $\hat{\bfS}^{(l)}_{ij} = \alpha_{ij}$. %if there exists an edge $(i,j) \in \mathcal{E}$ and 0 otherwise.
%Following works like superGAT \cite{kim2021howSuperGAT} and GATv2 \cite{brody2022how_GATV2} propose modifications to  \eqref{eq:attentionCoefficients} to improve the flexibility and expressiveness of GAT.
%We note that our proposition of $\omega$GNN, presented in Sec. \ref{sec:general_omega_gnn} is relevant to all of the mentioned methods, and more generally to any GNN the conforms to the form of Eq. \eqref{eq:general_gnn}.

%We define $\omega$GAT by arranging the learnt edge-weights $\alpha^{(l)}_{ij}$ from  \eqref{eq:attentionCoefficients} into a matrix $\hat{\bfS}^{(l)} \in \mathbb{R}^{n\times n}$ such that $\hat{\bfS}_{ij}^{(l)} = \alpha^{(l)}_{ij}$. 
To avoid repeated equations, we skip the per-layer $\omega$ formulation (as in \eqref{eq:omegaGNN}) and directly define the per-channel $\omega$GAT as follows
\begin{equation}
    \label{eq:omegaGAT}
    \bff^{(l+1)} = \sigma\left(\left(\bfI - \boldsymbol{\Omega}_{\vec{\omega}^{(l)}} \left(\bfI - \hat{\bfS}^{(l)}\right)\right)\bff^{(l)}\bfK^{(l)}\right).
\end{equation}
The introduction of $\boldsymbol{\Omega}_{\vec{\omega}^{(l)}}$ yields  a learnable propagation operator per layer and channel. We note that it is also possible to obtain multiple propagation operators from GAT by using a multi-head attention. %That is, in the case of a $H$ multi-head attention we follow the same formulation as in \cite{velickovic2018graph}, where multiple spatial operators are obtained using a set of learnt vectors $\textstyle{\{\bfa_{h}^{(l)}\}_{h=0}^{H-1}}$, and hence multiple operators $\bfS^{(l)}_h$ for each GAT layer are constructed analogously.
However, we distinguish our proposition from GAT in a 2-fold fashion. First, our propagation operators belong to a broader family that includes smoothing and sharpening operators as opposed to smoothing-only due to the SoftMax normalization in GAT. Secondly, our method requires less computational overhead when adding more propagation operators, as our $\omega$GAT requires a scalar per operator, while GAT doubles the number of channels to obtain more attention-heads. Also, utilizing a multi-head GAT can still lead to over-smoothing, as all the heads induce a non-negative operator.

%While GAT does not use a pre-determined propagation operator as assumed in Theorem. \ref{theorem:single_omega}, it still imposes strictly non-negative weights due to the use of SoftMax normalization. Another property of the propagation operator $\bfS^{(l)}_{ij} = \alpha_{ij}$, that stems from  \eqref{eq:attentionCoefficients} is that $\bfS^{(l)}$ is row-stochastic, and therefore it holds that $\bfS^{(l)} \mathbf{1} = \mathbf{1}$. This is an important observation since by the Perron-Frobenius theorem, it can be concluded that assuming that the matrix is irreducible, any $\bfS^{(l)}$ parameterized by GAT has a leading eigenvector of $\mathbf{1}$ and an eigenvalue of 1 \cite{horn2012matrix}. Therefore, the application of a series of propagation operators $\bfS^{(0)} , \ldots , \bfS^{(L-1)}$, even if they are different than each other, may converge to a constant vector, which makes GAT prone to over-smoothing.

To study the behaviour of our $\omega$GAT, we inspect its node features energy compared to GAT. %show that GAT over-smooths while our $\omega$GAT does not by inspecting the node features energy. We note that the over-smoothing of GAT was already discussed in \cite{Zhao2020PairNorm:, measuringoversmoothing}.
To this end, we define the GAT energy as 
\begin{equation}
    \label{eq:gatEnergy}
    E_{\rm{GAT}}(\bff^{(l)}) =\sum_{i \in \mathcal{V}} \sum_{j \in \mathcal{N}_i} \frac{1}{2} ||\bff^{(l)}_i - \bff^{(l)}_j ||_2^2. %= \frac{1}{2}\|\bfG\bff^{(l)}\|_2^2.
\end{equation}
This modification of the Dirichlet energy from \eqref{eq:dirichletEnergy} is required because in GAT \citep{velickovic2018graph} the leading eigenvector of the propagation operator $\hat{\bfS}^{(l)}$ is the constant vector $\textbf{1}$ as shown by \citet{measuringoversmoothing}, unlike the vector $\tilde\bfD^{\frac{1}{2}}\mathbf{1}$ in the symmetric normalized $\tilde{\bfP}$ from GCN \citep{kipf2016semi} where the Dirichlet energy is natural to consider \citep{Pei2020Geom-GCN:}.

We present the energy of a 64 layer GAT trained on the Cora dataset in Fig. \ref{fig:energyFig1}. It is evident that the accuracy degradation of GAT reported by \cite{Zhao2020PairNorm:} is in congruence with the decaying energy in \eqref{eq:gatEnergy}, while our $\omega$GAT does not experience decaying energy nor accuray degradation as more layers are added, as can be seen in Tab. \ref{table:semisupervised}. To further validate our findings, we repeat this experiment in Appendix \ref{appendix:gat_oversmoothing} on additional datasets and reach the same conclusion.

%From Theorem \ref{theorem:smoothing}, we can understand \emph{why} in addition to GCN \cite{kipf2016semi}, which was already proved to be smoothing \cite{wu2019simplifying}, methods that rely on GAT also over-smooth - the use of the Softmax normalization guarantees a spatial operator with non-negative entries, and as we present in Fig. \ref{fig:coraGAT_omega}, empirically, the learnt spatial operators by GAT will have a spectral radius that is smaller than 1, and thus will inevitably cause smoothing, making it prone to over-smoothing. \cite{mei2020escaping} \tred{maybe this paper is in order here ? they also talk about how using softmax can cause problems in learning}

\subsection{Computational Costs}
\label{sec:computationalcost}
Our $\omega$GNN approach is general and can be applied to any GNN that conforms to the structure of  \eqref{eq:general_gnn} and can be modified into  \eqref{eq:omegaGNN_perChannel}. The additional parameters compared to the baseline GNN are the added $\boldsymbol{\Omega}_{\vec{\omega}^{(l)}} \in \mathbb{R}^{c}$ parameters at each layer, yielding a relatively low computational overhead. For example, in GCN \citep{kipf2016semi} there are $c \times c$ trainable parameters requiring $c \times c \times n$ multiplications due to the $1\times1$ convolution $\bfK^{(l)}$. In our $\omega$GCN, we will have $c \times c + c$ parameters and $(c+1) \times c \times n$ multiplications. That is in addition to applying the propagation operators $\bfS^{(l)}$, which are identical for both methods. A similar analysis holds for GAT. To validate the actual complexity of our method, we present  the training and inference times for $\omega$GCN and $\omega$GAT in Appendix \ref{appendix:runtimes}. We see a negligible addition to the runtimes compared to the baselines, at the return of better performance.

\iffalse
\subsection{Implementation details}
In our experiments in Sec. \ref{sec:experiments} we use two architectures, one is similar to GCN \cite{kipf2016semi} (for node classification tasks) and the other is similar to GIN \cite{xu2018how} (for the graph classification task). We specify the exact architecture in Appendix \ref{sec:appendix_architectures}. In all architectures, in the first embedding layer the input node features ${\bff_{in}}$ are fed through a ($1\times 1$ convolution) layer ${\bf K}_{o}$ to obtain the initial features of our $\omega$GNN network:
${\bf f}^{(0)} = {\bf K}_o {\bff_{in}}$.  At the final layer of our network, we project ${\bf f}^{(L)}$ by 
${\bff}_{out} = {\bf K}_c {\bf f}^{(L)}$.  
Here ${\bf K}_c$ is a $1\times 1$ convolution layer mapping the hidden feature space to the output shape.

We initialize the embedding and projection layers with the Glorot \cite{glorot2010understanding} initialization, and $\bfK^{(l)}$ from  \eqref{eq:omegaGNN} is initialized with an identity matrix of shape $c \times c$. The initialization of $\Omega^{(l)}$ also starts from a vectors of ones. We note that our initialization yields a standard smoothing process, which is then adapted to the data as the learning process progresses, and if needed also changes the process to a non-smoothing one by the means of mixed-signs, as discussed earlier and specifically in Theorem. \ref{theorem:omega_smoothing}.
\fi

\section{Other Related Work} \label{sec:related}

\textbf{Over-smoothing in GNNs.} The over-smoothing phenomenon was identified by \cite{li2018deeper}, and was profoundly studied in recent years. Various methods stemming from different approaches were proposed. For example, methods like DropEdge \citep{Rong2020DropEdge:}, PairNorm \citep{Zhao2020PairNorm:}, and EGNN \citep{zhou2021dirichlet} propose augmentation, normalization and energy-based penalty methods to alleviate over-smoothing, respectively. Other methods like \cite{min2020scattering} propose to augment GCN with geometric scattering transforms and residual convolutions, and GCNII \citep{chen20simple} present a spectral analysis of the smoothing property of GCN \citep{kipf2016semi} and propose adding an initial identity residual connection and a decay of the weights of deeper layers, which are also used in EGNN \citep{zhou2021dirichlet}. 

\textbf{Graph Neural Diffusion.}
The view of GNNs as a diffusion process has gained popularity in recent years. Methods like APPNP \citep{klicpera2018combining} propose to use a personalized PageRank \citep{page1999pagerank} algorithm to determine the diffusion of features, and GDC \citep{gasteiger_diffusion_2019} imposes constraints on the ChebNet \citep{defferrard2016convolutional} architecture to obtain diffusion kernels, showing accuracy improvement. Other works like GRAND \citep{chamberlain2021grand}, CFD-GCN \citep{belbute_peres_cfdgcn_2020}, PDE-GCN \citep{eliasof2021pde} and GRAND++ \citep{thorpe2022grand} propose to view GNN layers as time steps in the integration process of ODEs and PDEs that arise from a non-linear heat equation, allowing to control the  diffusion (smoothing) in the network to prevent over-smoothing. In addition, some GNNs \citep{eliasof2021pde,rusch2022graph} propose a mixture between diffusion and oscillatory processes to avoid over-smoothing by    frequency preservation of the features.

\textbf{Mixed-sign operators in GNNs.}
The importance of mixed-sign high-pass vs low-pass filters was discussed in  the spectral GPRGNN \citep{chien2021adaptive}, which utilizes a single high-order polynomial (k-hop) filter. The authors show that weights corresponding to high-pass filters are obtained for heterophilic datasets. \tred{In \cref{appendix:learntOmega}, we show similar conclusions.} Other similar spectral methods include JacobiConv \citep{JacobiConv2022}, BernNet \citep{he2021bernnet}. In all these methods, a high-order polynomial is learnt, where all the propagation layers are stacked linearly one after the other. In contrast, we wish to follow the non-linear form \eqref{eq:general_gnn} with a 1-hop propagation and $1\times1$ convolution for each layer, as this is a form that is closest to standard CNNs, which have been proven to be effective for many challenging tasks in computer vision \cite{chen2017deeplab, tan2019efficientnet, sandler2018mobilenetv2}. \tred{Mixed-sign operators were also presented in the works of  \citet{yang2021diverse} and \citet{yan2021two}, employing attention-based propagation operator with a tanh/cosine activation function to obtain values in $[-1,1]$. This mechanism is different than our weighting approach which in which the type of operation is learnt directly. } Lastly, mixed-sign operators were also discussed in \cite{eliasof2022pathgcn}, where $k$-hop filters and stochastic path sampling mechanisms are utilized. However, such a method requires significantly more computational resources than a standard GNN like \cite{kipf2016semi} due to the path sampling strategy and larger filters of $5$-hop required for optimal accuracy. However, our $\omega$GNNs perform 1-hop propagations and, as we show in Appendix \ref{appendix:runtimes}, obtain competitive results without significant added computational costs.

\section{Experiments}
\label{sec:experiments}

\begin{table*}[t]
  \caption{Summary of semi-supervised node classification accuracy (\%)}
  \label{table:semisupervised_summary}
  \begin{center}
  \begin{tabular}{lccccccccccc}
  \toprule
    Method  & GCN & GAT & APPNP  & GCNII & GRAND & superGAT & EGNN & $\omega$GCN (Ours) & $\omega$GAT (Ours)     \\
    \midrule
    Cora & 81.1 & 83.1 & 83.3 &  85.5 & 84.7 & 84.3 & 85.7 & \textbf{85.9} & 84.8 \\ 
    Citeseer &  70.8 & 70.8 & 71.8  & 73.4 & 73.6 & 72.6 & -- & 73.3 & \textbf{74.0} \\
    Pubmed   & 79.0 & 78.5 & 80.1  & 80.3 & 71.0 & 81.7  & 80.1 & 81.1 & \textbf{81.8} \\
    \bottomrule
  \end{tabular}
\end{center}
\end{table*}

%%%%%%%%%%%%%%%%

We demonstrate our $\omega$GCN and $\omega$GAT on node classification, inductive learning and graph classification tasks. Additionally, we conduct an ablation study of the different configurations of our method and experimentally verify the theorems from Sec. \ref{sec:wgnns}.
A description of the network architectures is given in Appendix \ref{appendix:architectures}. We use the Adam \citep{kingma2014adam} optimizer in all experiments, and perform grid search to determine the hyper-parameters reported in Appendix \ref{appendix:hyperparameters}. The objective function in all experiments is the cross-entropy loss, besides inductive learning on PPI \citep{hamilton2017inductive} where we use the binary cross-entropy loss. Our code is implemented with PyTorch \citep{pytorch} and PyTorch-Geometric \citep{pyg2019} and trained on an Nvidia Titan RTX GPU.

We show that for all the considered tasks and datasets, whose statistics are provided Appendix \ref{appendix:datasets}, our $\omega$GCN and $\omega$GAT are either better or on par with other state-of-the-art models.

%%%%%%%%%%%%%%%%

%%%%%%%%%%%%%%%%%%%%%%%%%%%%
\subsection{Node Classification}
\label{sec:semiSupervised_experiment}
We employ the Cora, Citeseer and Pubmed \citep{sen2008collective} datasets using the standard training/validation/testing split by \cite{yang2016revisiting}, with 20 nodes per class for training, 500 validation nodes and 1,000 testing nodes. We follow the training and evaluation scheme of \cite{chen20simple} and compare with models like GCN, GAT, superGAT \citep{kim2021howSuperGAT}, Inception \citep{szegedy2017inception}, APPNP \citep{klicpera2018combining}, JKNet \cite{jknet}, DropEdge \citep{Rong2020DropEdge:}, GCNII \citep{chen20simple}, GRAND \citep{chamberlain2021grand}, PDE-GCN \citep{eliasof2021pde} and EGNN \citep{zhou2021dirichlet}. We summarize the results in Tab. \ref{table:semisupervised_summary} where we see better or on par performance with other existing methods. %For example, we obtain $85.9 \%$ accuracy on Cora using our $\omega$GCN compared to $85.5 \%$ and $85.7 \%$ with GCNII and EGNN, respectively.
Additionally, we report the accuracy per number of layers, from 2 to 64 in Tab. \ref{table:semisupervised}, where it is evident that our $\omega$GCN and $\omega$GAT do not over-smooth. To ensure the robustness of our method, we also experiment with 100 random splits in Appendix \ref{appendix:StatisticalSig} where our $\omega$GCN and $\omega$GAT perform similarly to the splits in Table \ref{table:semisupervised_summary}.

%%%%%%%%%%%%%%%%%%%%%%%%%%%

\begin{table*}[t]
  \caption{Semi-supervised node classification accuracy ($ \%$). -- indicates not available results.}
  \label{table:semisupervised}
  \begin{center}
  %\begin{small}
  \resizebox{1.0\linewidth}{!}{\setlength{\tabcolsep}{1.mm}{
  \begin{tabular}{l|cccccc|cccccc|cccccc}
    \toprule
    %&\multicolumn{7}{c}{Layers}\\
    %\cline{3-8}
    %Dataset & Method & 2 & 4 & 8 & 16 & 32 & 64 \\
   \multirow{1}{*}{Dataset} & \multicolumn{6}{c|}{Cora} & \multicolumn{6}{c|}{Citeseer} & \multicolumn{6}{c}{Pubmed} \\ 
                         Layers & 2  & 4  & 8 & 16 & 32 & 64 & 2  & 4  & 8 & 16 & 32 & 64 & 2  & 4  & 8 & 16 & 32 & 64 \\
    \midrule
     GCN & \bf{81.1} & 80.4 & 69.5 & 64.9 & 60.3 & 28.7 & \bf{70.8} & 67.6 & 30.2 & 18.3 & 25.0 & 20.0 & \bf{79.0} & 76.5 & 61.2 & 40.9 & 22.4 & 35.3 \\
    GCN (Drop) & \bf{82.8} & 82.0 & 75.8 & 75.7 & 62.5 & 49.5 & \bf{72.3} & 70.6 & 61.4 & 57.2 & 41.6 & 34.4 & \bf{79.6} & 79.4 & 78.1 & 78.5 & 77.0 & 61.5 \\
    JKNet  & -- & 80.2 & 80.7 & 80.2 & \bf{81.1} & 71.5 & -- & 68.7 & 67.7 & \bf{69.8} & 68.2 & 63.4 & -- & 78.0 & \bf{78.1} & 72.6 & 72.4 & 74.5 \\
    JKNet (Drop) & -- & \bf{83.3} & 82.6 & 83.0 & 82.5 & 83.2 & -- & 72.6 & 71.8 & \bf{72.6} & 70.8 & 72.2 & -- & 78.7 & 78.7 & \bf{79.7} & 79.2 & 78.9 \\
    Incep & -- & 77.6 & 76.5 & 81.7 & \bf{81.7} & 80.0 & -- & 69.3 & 68.4 & \bf{70.2} & 68.0 & 67.5 & -- & 77.7 & \bf{77.9} & 74.9 & -- & --   \\
    Incep (Drop)  & -- & 82.9 & 82.5 & 83.1 & 83.1 & \bf{83.5} & -- & \bf{72.7} & 71.4 & 72.5 & 72.6 & 71.0 & -- & \bf{79.5} & 78.6 & 79.0 & -- & --  \\
    GCNII  & 82.2 & 82.6 & 84.2 & 84.6 & 85.4 & \bf{85.5} & 68.2 & 68.8 & 70.6 & 72.9 & \bf{73.4} & 73.4 & 78.2 & 78.8 & 79.3 & \bf{80.2} & 79.8 & 79.7 \\
    GCNII*& 80.2 & 82.3 & 82.8 & 83.5 & 84.9 & \bf{85.3} & 66.1 & 66.7 & 70.6 & 72.0 & \bf{73.2} & 73.1  & 77.7 & 78.2 & 78.8 & \bf{80.3}& 79.8 & 80.1 \\
    PDE-GCN\textsubscript{D} & 82.0 & 83.6 & 84.0 & 84.2 & 84.3 & \bf{84.3} & 74.6 & 75.0 & 75.2 & 75.5 & \textbf{75.6} & 75.5 & 79.3 & \bf{80.6} & 80.1  & 80.4 & 80.2 & 80.3  \\
    EGNN & 83.2 & -- & -- & 85.4 & -- & \textbf{85.7} & -- & -- & -- & -- & -- & -- & 79.2 & -- & -- & 80.0 & -- & \textbf{80.1}  \\
    $\omega$GCN (Ours) & 82.6 & 83.8 & 84.2 & 84.4 & 85.5 & \textbf{85.9} & 71.3 & 71.6 & 72.1 & 72.4 & 73.3 & \textbf{73.3}  & 79.7 & 80.2 & 80.1 & 80.5 & 80.8 & \textbf{81.1}  \\
    $\omega$GAT (Ours) &  83.4 & 83.7 & 84.0 & 84.3 & 84.4 & \textbf{84.8} & 72.5 & 73.1 & 73.3 & 73.5 & 73.9 & \textbf{74.0} & 80.3 & 81.0 &  81.2 & 81.3 & 81.5 & \textbf{81.8}  \\
    \bottomrule
  \end{tabular}}}
  %\end{small}
\end{center}
\end{table*}

\begin{table}[]
\begin{minipage}[t]{1\linewidth}
  \caption{Node classification accuracy ($ \%$) on \emph{homophilic} datasets. $\dagger$ denotes the maximal accuracy of several proposed variants.}
  \label{table:homophilic_fully}   
  \center{
    \footnotesize
  \begin{tabular}{lccc}
    \toprule
    Method & Cora & Citeseer & Pubmed \\
    Homophily & 0.81 & 0.80 & 0.74 \\
    \midrule
    GCN  & 85.77 & 73.68 & 88.13  \\
    GAT & 86.37 & 74.32 & 87.62 \\
    GCNII\textsuperscript{$\dagger$} & 88.49  & 77.13 & 90.30  \\
    Geom-GCN\textsuperscript{$\dagger$} & 85.27 & 77.99 & 90.05\\
    %Geom-GCN-P & 84.93 & 75.14 & 88.09\\
    %Geom-GCN-S & 85.27 & 74.71 & 84.75\\
    APPNP &  87.87 & 76.53 & 89.40 \\
    JKNet & 85.25  & 75.85  & 88.94 \\
    %JKNet (Drop) & 87.46  & 75.96 & 89.45 \\
    WRGAT & 88.20 & 76.81 & 88.52 \\
    %Incep (Drop) & 86.86 & 76.83 & 89.18 \\
    %GCNII*  & 88.01 & 77.13 & 90.30\\
    PDE-GCN\textsubscript{M} & 88.60  & \textbf{78.48} & 89.93 \\
    NSD\textsuperscript{$\dagger$}  & 87.14  & 77.14   & 89.49 \\
    %O(d)-NSD  & 86.90  & 76.70   & 89.49 \\
    GGCN  & 87.95  & 77.14   & 89.15 \\
    H2GCN  & 87.87  & 77.11   & 89.49 \\
    C\&S & 89.77 & 77.29 & 90.01\\
    DMP\textsuperscript{$\dagger$}  & 86.52 & 76.87 & 89.27 \\
    LINKX  & 84.64 & 73.19 & 87.86 \\ 
    ACMII-GCN++ & 88.25 & 77.12 & 89.71 \\
    %pathGCN &  \textbf{90.02} (64) &  \textbf{78.95} (32) &  \textbf{90.42} (64) &  \textbf{66.79} (16) &  \textbf{91.35} (8)  &  \textbf{95.14} (16) &  \textbf{93.53} (16) \\
    \midrule
    $\omega$GCN (Ours) &  \textbf{89.30}  & 77.88  &  90.45 
   \\
    $\omega$GAT (Ours) &  89.25  & 78.01  & \textbf{90.65}
   \\

    \bottomrule
  \end{tabular}}
  \end{minipage}
\end{table}

\begin{table}[]
\footnotesize
      \begin{minipage}[t]{1\linewidth}
\centering
  \caption{Node classification accuracy ($ \%$) on \emph{heterophilic} datasets. $*$ denotes the maximal accuracy of several proposed variants.}
  \label{table:heterophilic_fully}
   
  \begin{center}
  {\setlength{\tabcolsep}{1.3mm}{
  \begin{tabular}{lcccccc}
    \toprule
    Method & Squirrel & Film &  Cham. & Corn. & Texas & Wisc. \\
    Homophily & 0.22 & 0.22 & 0.23 & 0.30  & 0.11 & 0.21 \\
        \midrule
    GCN  & 23.96 & 26.86 &  28.18 &  52.70 & 52.16 & 48.92 \\
    GAT & 30.03 & 28.45 & 42.93 & 54.32 & 58.38 & 49.41
    \\
    GCNII & 38.47 & 32.87 &   60.61  & 74.86 & 69.46 & 74.12 \\
    Geom-GCN\textsuperscript{$*$}& 38.32 & 31.63 &  60.90 & 60.81 & 67.57 & 64.12 \\
    %Geom-GCN-P & 38.14 & 31.63 & 60.90 & 60.81  & 67.57 & 64.12 \\
    %Geom-GCN-S & 36.24 & 30.30 &   59.96 & 55.68  & 59.73 & 56.67 \\
    %APPNP &   54.30 & 73.51  & 65.41 & 69.02 \\
    %JKNet & 60.07 & 57.30 & 56.49 & 48.82  \\
    %JKNet (Drop) & & &  62.08  & 61.08 & 57.30 & 50.59  \\
    MixHop & 43.80 & 32.22 & 60.50 & 73.51 & 77.84 & 75.88 \\ 
    %Incep (Drop) &   61.71 & 61.62 & 57.84 & 50.20 \\
    %GCNII*  & & & 62.48 & 76.49 & 77.84  & 81.57  \\
    PDE-GCN\textsubscript{M} & -- & -- &   66.01 & 89.73  & 93.24  &  91.76 \\
    GRAND & 40.05 & 35.62 &  54.67 & 82.16 & 75.68 & 79.41 \\ 
    NSD\textsuperscript{$*$}  &  56.34 & 37.79  & 68.68  &  86.49 & 85.95 & 89.41 \\
    %O(d)-NSD  &  56.34 & 37.81 & 68.04 & 84.86 & 85.95 & 89.41 \\
    WRGAT & 48.85 & 36.53 & 65.24 &  81.62 & 83.62 & 86.98 \\
    MagNet &  --  & -- & --  & 84.30 & 83.30 & 85.70 \\ 
    GGCN  & 55.17 & 37.81 &  71.14 & 85.68  & 84.86  &  86.86 \\
    H2GCN  & 36.48 & 35.70 & 60.11 & 82.70  & 84.86  &  87.65 \\
    GraphCON\textsuperscript{$*$} & -- & -- & -- & 84.30 & 85.40 & 87.80 \\  
    %GraphCON-GAT & --  & --  & -- & 83.20 & 82.20 & 85.70 \\  
    FAGCN & 42.59 & 34.87 & 55.22 & 79.19 & 82.43 & 82.94 \\
    GPRGNN & 31.61 & 34.63 & 46.58 & 80.27 & 78.38 & 82.94 \\
    DMP\textsuperscript{$*$}  & 47.26 & 35.72 & 62.28 & 89.19 & 89.19 & 92.16 \\
    ACMP-GCN  & --  & -- & -- & 85.4 & 86.2 & 86.1  \\ 
    LINKX  & 61.81 &  36.10 & 68.42 & 77.84 & 74.60 & 75.49 \\
    G\textsuperscript{2}\textsuperscript{$*$} & 64.26 &  37.30 & 71.40 & 87.30 &  87.57 & 87.84\\
    ACMII-GCN++ & \textbf{67.40} & 37.09 & \textbf{74.76} & 86.49 & 88.38 & 88.43 \\
    %pathGCN &  \textbf{90.02} (64) &  \textbf{78.95} (32) &  \textbf{90.42} (64) &  \textbf{66.79} (16) &  \textbf{91.35} (8)  &  \textbf{95.14} (16) &  \textbf{93.53} (16) \\
    \midrule    
    %DRGNN-T  (Ours)& EXPLAIN WHY NOT APPLICABLE\\ 
        $\omega$GCN (Ours) & 59.41 & \textbf{38.94}  & 70.02  & 91.35  & 94.05   & 92.35 
   \\
    $\omega$GAT (Ours) & 58.96 & 38.64 &72.23 & \textbf{91.62}  & \textbf{94.59} & \textbf{92.94}
   \\
    \bottomrule
  \end{tabular}
  }}
\end{center}
\end{minipage}
\end{table}

  \begin{table}
    \caption{Inductive learning on PPI dataset. Results are reported in micro-averaged F1 score.}
  \label{table:ppi}
  \begin{center}
  \resizebox{\linewidth}{!}{
  \begin{tabular}{lcc}
    \toprule
    Method & Micro-averaged F1 \\
        \midrule
    GCN \cite{kipf2016semi} & 60.73  \\
    GraphSAGE \cite{hamilton2017inductive} & $61.20$ \\
    VR-GCN \citep{vrgcn} & 97.80 \\
    GaAN \citep{zhang18} & 98.71 \\
    GAT \citep{velickovic2018graph} &  97.30 \\
    JKNet \citep{jknet} & 97.60 \\
    GeniePath \citep{geniepath} & 98.50 \\
    Cluster-GCN \citep{clustergcn} & 99.36 \\
    %GCNII \cite{chen20simple} & 99.54 \\
    GCNII* \citep{chen20simple} & 99.58 \\
    PDE-GCN\textsubscript{M} \citep{eliasof2021pde} & 99.18  \\
    \midrule
    $\omega$GCN (Ours) & \textbf{99.60}  \\
    $\omega$GAT (Ours) & 99.48  \\
    \bottomrule
  \end{tabular}}
\end{center}
%}
\end{table}

To further validate the efficacy of our method on fully-supervised node classification, both on homophilic and heterophilic datasets as defined in \cite{Pei2020Geom-GCN:}. Specifically, examine our $\omega$GCN and $\omega$GAT on Cora, Citeseer, Pubmed, Chameleon \citep{musae}, Film, Cornell, Texas and Wisconsin using the identical train/validation/test splits of $48 \%, 32\%, 20\%$, respectively, and report the average performance over 10 random splits from \cite{Pei2020Geom-GCN:}. %In all experiments, 64 channels are used and a grid search is performed to determine the hyper-parameters.
We compare our performance with, GCN, GAT, Geom-GCN, APPNP, JKNet, Inception, GCNII, PDE-GCN  and others, as presented in Tab. \ref{table:homophilic_fully}-\ref{table:heterophilic_fully}.  Additionally, we evaluate our $\omega$GCN and $\omega$GAT on the Ogbn-arxiv \citep{hu2020ogb} dataset, as reported in Tab. \ref{table:arxiv} in the Appendix.
We see an accuracy improvement across all benchmarks compared to the considered methods. In Appendix \ref{appendix:learntOmega} we present and discuss the learnt $\vec{\omega}$ for homophilic and heterophilic datasets. 

%For example, our $\omega$GCN achieves $89.30\%$ accuracy on Cora with our $\omega$GCN, compared to $88.49\%$ with GCNII and $88.60\%$ with PDE-GCN.

\subsection{Inductive Learning}
\label{sec:inductivelearningppi}
We employ the PPI dataset \citep{hamilton2017inductive} for the inductive learning task. We use 8 layer $\omega$GCN and $\omega$GAT, with a learning rate of 0.001, dropout of 0.2 and no weight-decay. As a comparison we consider several methods and report the micro-averaged F1 score in in Tab. \ref{table:ppi}. Our $\omega$GCN achieves a score of $99.60$, which is higher than the other methods like GAT, JKNet, GeniePath, Cluster-GCN and PDE-GCN.

\subsection{Graph Classification}
\label{sec:TUDatasets_results}
Previous experiments considered the \emph{node-classification} task. To further demonstrate the efficacy of our $\omega$GNNs we experiment with graph classification on  TUDatasets \citep{Morris2020TUDatasets}. Here, we follow the same experimental settings from \cite{xu2018how}, and report the 10 fold cross-validation performance on MUTAG, PTC, PROTEINS, NCI1 and NCI109 datasets. The hyper-parameters are determined by a grid search, as in \cite{xu2018how} and are reported in Appendix E. We compare our $\omega$GCN and $\omega$GAT with recent and popular methods like GIN \citep{xu2018how}, GCONV \citep{morris2019weisfeiler}, RNI \cite{abboud2020surprising}, DGCNN \citep{zhang2018end}, IGN \citep{maron2018invariant}, GSN \citep{bouritsas2022improving}, SIN \citep{bodnar2021weisfeiler}, CIN \citep{bodnar2021CW} and others. We also compare with methods that stem from 'classical' graph algorithms like PK \citep{neumann2016propagation} and WL Kernel \citep{shervashidze2011weisfeiler}.
All  the results are summarized in Tab. \ref{table:graphhclassification_results}, with an evident improvement or similar results to current deep learning as well as classical methods, highlighting the efficacy of our approach.

\begin{table}[t]
  \caption{Graph classification accuracy ($\%$) on TUDatasets \citep{Morris2020TUDatasets}.}
  \label{table:graphhclassification_results}
  \begin{center}
  \resizebox{1.0\linewidth}{!}{\begin{tabular}{lccccc}
    \toprule
    Method & MUTAG & PTC & PROTEINS & NCI1 & NCI109 \\
        \midrule
    %RWK \citep{gartner2003graph} & 79.2 $\pm$ 2.1 &  55.9  $\pm$  0.3 &  59.6  $\pm$  0.1 &  -- & -- \\
    %GK \citep{shervashidze2009efficient} & 81.4 $\pm$ 1.7 &  55.7 $\pm$ 0.5 &  71.4 $\pm$ 0.3 &  62.5 $\pm$ 0.3 &  62.4$\pm$0.3  \\
    PK  & 76.0 $\pm$  2.7  & 59.5$\pm$ 2.4 &  73.7$\pm$ 0.7 &  82.5$\pm$ 0.5 &  -- \\
    WL Kernel & 90.4$\pm$ 5.7 &  59.9$\pm$ 4.3 &  75.0$\pm$ 3.1 &  \textbf{86.0$\pm$ 1.8} &  -- \\
    \midrule
    DGCNN  & 85.8$\pm$1.8 &  58.6$\pm$2.5 &  75.5$\pm$0.9 &  74.4$\pm$0.5 &  -- \\
    IGN   & 83.9$\pm$13.0 &  58.5$\pm$6.9 &  76.6$\pm$5.5 &  74.3$\pm$2.7 &  72.8$\pm$1.5 \\
    PPGNS & 90.6$\pm$8.7 &  66.2$\pm$6.6 &  77.2$\pm$4.7 &  83.2$\pm$1.1 &  82.2$\pm$1.4 \\
    %NATURAL GN \citep{de2020natural} & 89.4$\pm$1.6 &  66.8$\pm$1.7 &  71.7$\pm$1.0 &  82.4$\pm$1.3 &  -- \\
    GSN  & 92.2$\pm$7.5 &  68.2$\pm$7.2 &  76.6$\pm$5.0 &  83.5$\pm$2.0 &  -- \\
    SIN  & -- &  -- &  76.4$\pm$3.3 &  82.7$\pm$2.1 & -- \\
    CIN & 92.7$\pm$3.6 &  68.2$\pm$3.5 &  77.0$\pm$3.4 &  83.6$\pm$3.1 &  84.0$\pm$3.1 \\
    %\midrule
    GIN  & 89.4$\pm$5.6 &  64.6$\pm$7.0 &  76.2$\pm$2.8 &  82.7$\pm$1.7 &  82.2$\pm$1.6 \\
    %\midrule
    %GIN + ID \citep{you2021identity}  & 90.4$\pm$5.4 &  67.2$\pm$4.3 &  75.4$\pm$2.7 &  82.6$\pm$1.6 &  82.1$\pm$1.5 \\
    %\midrule
    %DROP \citep{Rong2020DropEdge:} & 91.0$\pm$5.7 &  64.5$\pm$2.6 &  73.5$\pm$4.5 &  82.0$\pm$2.6 & 82.2$\pm$1.4 \\
    %\midrule
    GCONV  & 90.5$\pm$4.6 &  64.9$\pm$10.4 &  73.9$\pm$6.1 &  82.4$\pm$2.7 &  81.7$\pm$1.0 \\
    %\midrule
    RNI  & 91.0$\pm$4.9 &  64.3$\pm$6.1 &  73.3$\pm$3.3 &  82.1$\pm$1.7 &  81.7$\pm$1.0 \\
    
    \midrule
    $\omega$GCN (Ours) &  94.6 $\pm$ 4.1 & 73.8 $\pm$ 4.3 & 80.2 $\pm$ 2.5 & 84.1 $\pm$ 1.2 & \textbf{84.5 $\pm$ 1.8}  \\
    $\omega$GAT (Ours) &  \textbf{95.2 $\pm$  3.7} & \textbf{75.8  $\pm$ 3.5}  & \textbf{80.7 $\pm$ 3.7}  & 84.4  $\pm$ 1.7  &  83.6 $\pm$ 1.2   \\
    \bottomrule
  \end{tabular}}
\end{center}
\end{table}

\subsection{Ablation Study}
\label{sec:ablation}
In this section we study the different components and configurations of our $\omega$GNN. We start by allowing a global (single) $\omega$ to be learnt throughout all the layers---this architecture is dubbed as $\omega$GCN\textsubscript{G}. We validate that this simple variant does not over-smooth, depicted in Tab. \ref{table:ablationVariants}. The table also shows $\omega$GCN\textsubscript{PL}, that includes a single parameter $\omega^{(l)}$ per layer, and $\omega$GCN shown in the results earlier that has $\Omega^{(l)}$, i.e., a parameter per layer and channel, which yields further accuracy improvements. In addition, we empirically verify our theoretical results from Sec. \ref{sec:wgnns} in Fig. \ref{fig:sumOmega}, where we show that the obtained values of $\omega$ (whether the global or averaged per-layer ones)
scale as $1/L$ and behave according to Theorem \ref{theorem:single_omega} and Corollary \ref{cor:sumOmega}.
%We also present the sum of the learnt $\omega^{(l)}$ for different datasets in Fig. \ref{fig:sumOmega} to empirically validate Corollary \ref{cor:sumOmega}. %\tred{The results show the clear advantage of having different $\omega$ per layer and per channel as opposed to less learnable $\omega$'s.}
For completeness, we also perform the ablation study on $\omega$GAT in Appendix \ref{appendix:ablation}.

\begin{table}[t]
%\begin{figure}
%\begin{minipage}{.48\linewidth}
\captionof{table}{Accuracy ($\%$) of variants of $\omega$GCN on semi-supervised classification.}
\centering
  %\caption{Accuracy ($\%$) of variants of $\omega$GCN on semi-supervised classification.}
  \label{table:ablationVariants}
  \begin{center}
  \resizebox{1.0\linewidth}{!}{\setlength{\tabcolsep}{2.mm}{\begin{tabular}{llcccccc}
    \toprule
    %&\multicolumn{7}{c}{Layers}\\
    %\cline{3-8}
    %Dataset & Method & 2 & 4 & 8 & 16 & 32 & 64 \\
    \multirow{2}{*}{Data.} & \multirow{2}{*}{Variant} & \multicolumn{6}{c}{Layers} \\
                         &  & 2  & 4  & 8 & 16 & 32 & 64 \\
    \midrule
    Cora & $\omega$GCN\textsubscript{G} & 83.4 &	84.3 &	84.2 &	84.1 &	84.3 &	84.4
     \\
    & $\omega$GCN\textsubscript{PL} & 83.0 & 83.6 & 84.0 & 84.2 & 84.5 & 84.8    \\
    & $\omega$GCN &  82.6 & 83.8 & 84.2 & 84.4 & 85.5 & 85.9   \\
   \midrule
    Cite. & $\omega$GCN\textsubscript{G} & 71.0 &	71.4 &71.3 & 71.7 &	72.0 &	71.8
   \\
    & $\omega$GCN\textsubscript{PL} & 71.1 & 71.3 & 71.5 & 71.8 & 72.4 & 72.6   \\
    & $\omega$GCN & 71.3 & 71.6 & 72.1 & 72.4 & 73.3 & 73.3\\
    \midrule
        Pub. & $\omega$GCN\textsubscript{G} & 79.8 & 80.4 & 80.5 & 80.4 & 80.2 & 80.3  \\
    & $\omega$GCN\textsubscript{PL} & 79.8 & 80.0 & 80.2 & 80.4 & 80.5 & 80.8   \\
    & $\omega$GCN & 79.7 & 80.2 & 80.1 & 80.5 & 80.8 & 81.1  \\

    \bottomrule
  \end{tabular}
  }}
\end{center}
\end{table}
%\end{minipage}\hfill
%\begin{minipage}{.49\linewidth}
\begin{figure}[t]
\begin{tikzpicture}
  \begin{axis}[
      width=1.0\linewidth, 
      height=0.8\linewidth,
      grid=major,
      ymode=log,
      grid style={dashed,gray!30},
      xlabel=Layers,
      ylabel=$\bar{\omega}$,
      ylabel near ticks,
      legend style={at={(0.65,0.999)},anchor=north,scale=1.0, draw=none, cells={anchor=west}, font=\tiny, fill=none},
      legend columns=3,
      xtick={0,1,2,3,4,5},
      xticklabels = {2,4,8,16,32,64},
      yticklabel style={
        /pgf/number format/fixed,
        /pgf/number format/precision=3
      },
      scaled y ticks=false,
      every axis plot post/.style={thick},
      ymax=1.1,
      ymin=0,
    ]
    \addplot[red, mark=oplus*, forget plot]
    table[x=layer,y=cora_avg,col sep=comma] {data/omega_sum.csv};
    \addplot[blue, mark=oplus*, forget plot]
    table[x=layer,y=citeseer_avg,col sep=comma] {data/omega_sum.csv};
    \addplot[green, mark=oplus*, forget plot]
    table[x=layer,y=pubmed_avg,col sep=comma] {data/omega_sum.csv};
    
    \addplot[red, style=dotted, mark=oplus*, forget plot]
    table[x=layer,y=avg_single_cora,col sep=comma] {data/omega_sum.csv};
        \addplot[blue, style=dotted, mark=oplus*, forget plot]
    table[x=layer,y=avg_single_citeseer,col sep=comma] {data/omega_sum.csv};
        \addplot[green, style=dotted, mark=oplus*, forget plot]
    table[x=layer,y=avg_single_pubmed,col sep=comma] {data/omega_sum.csv};

    %\addplot[magenta, style=dotted, mark=oplus*, forget plot]
    %table[x=layer,y=ref,col sep=comma] {data/omega_sum.csv};

    \addplot[red, draw=none] coordinates {(1,1)};
    \addplot[blue, draw=none] coordinates {(1,1)};
    \addplot[green, draw=none] coordinates {(1,1)};

    \addplot[gray, draw=none] coordinates {(1,1)};
    \addplot[gray, style=dotted, draw=none] coordinates {(1,1)};

    \legend{Cora, Citeseer, Pubmed, $\omega$GCN\textsubscript{PL}, $\omega$GCN\textsubscript{G}}

    \end{axis}
\end{tikzpicture}
%\caption{Total smoothing $\bar{\omega}$ vs. layers}
\captionof{figure}{\tred{The average value of  $\omega$ across the layers (denoted by $\bar{\omega} = \frac{1}{L}\sum_{l=0}^{L-1} \omega^{(l)}$) vs. the number layers for $\omega$GCN\textsubscript{G} and $\omega$GCN\textsubscript{PL}. %The slope of $\bar{\omega}$ in the plots is approximately the same as that of $\frac{1}{L}$. 
Here, $\bar \omega$ scales like $\frac{1}{L}$ for a varying $L$, in congruence with \cref{theorem:single_omega}.}}
\label{fig:sumOmega}
%\end{minipage}
\end{figure}

\section{Summary}
In this work we propose an effective and computationally light modification
% that applies to a large family of GNNs
of the large family of GNNs 
that carry the form of a separable propagation and $1\times1$ convolutions.
In particular we demonstrate its efficacy on the popular GCN and GAT architectures. Our theorems show that $\omega$GNNs can avoid over-smoothing 
as their learnable weighting factors $\vec{\omega}$  enable mixing smoothing and sharpening propagation operators.
This flexibility also enhances the expressiveness.
Through an extensive set of experiments on 15 datasets (ranging from node classification to graph classification), an ablation study, and comparisons to several recent methods, we validate our theoretical findings and demonstrate the performance of our $\omega$GNN.
% that validates our theoretical findings, we demonstrate the performance of our $\omega$GNN by comparing to several recent methods.
% , reading competitive performance compared to recent methods.
\label{sec:summary}

\section*{Acknowledgements}
The research reported in this paper was supported by grant no. 2018209 from the United States - Israel Binational Science Foundation (BSF), Jerusalem, Israel, and in part by the Israeli
Council for Higher Education (CHE) via the Data Science
Research Center, Ben-Gurion University of the Negev, Israel. ME is supported by Kreitman High-tech scholarship. LR’s work is also partially supported by NSF DMS 2038118, AFOSR grant FA9550-
20-1-0372, and US DOE Office of Advanced Scientific Computing Research Field Work Proposal 20-023231.


%\bibliography{example_paper}
\bibliography{main.bbl}
\bibliographystyle{icml2023}

%%%%%%%%%%%%%%%%%%%%%%%%%%%%%%%%%%%%%%%%%%%%%%%%%%%%%%%%%%%%%%%%%%%%%%%%%%%%%%%
%%%%%%%%%%%%%%%%%%%%%%%%%%%%%%%%%%%%%%%%%%%%%%%%%%%%%%%%%%%%%%%%%%%%%%%%%%%%%%%
% APPENDIX
%%%%%%%%%%%%%%%%%%%%%%%%%%%%%%%%%%%%%%%%%%%%%%%%%%%%%%%%%%%%%%%%%%%%%%%%%%%%%%%
%%%%%%%%%%%%%%%%%%%%%%%%%%%%%%%%%%%%%%%%%%%%%%%%%%%%%%%%%%%%%%%%%%%%%%%%%%%%%%%
\newpage
\appendix
\onecolumn

\appendix
\newtheorem{thm}{Theorem}
\newtheorem{cor}{Corollary}

\counterwithin*{thm}{subsection} 

\section{Proofs of Theorems}
\label{appendix:proofs}

Here we repeat the theorems, observations and corollaries from the main paper, for convenience, and provide their proofs or derivation.
\subsection*{$\tilde{\bfP}$ is a Scaled Diffusion Operator} 
%\sout{the Jacobi operator.}} 
Assume that $\bfA$ is the adjacency matrix, and $\bfD$ is the degree matrix.
Denote the adjacency matrix with added self-loops by $\tilde{\bfA} = \bfA + \bfI$. Then, the convolution operator from GCN \citep{kipf2016semi} is
\begin{equation}
    \label{appendixEq:gcn}
    \tilde\bfP = \tilde{\bfD}^{-\frac{1}{2}} \tilde{\bfA} \tilde{\bfD}^{-\frac{1}{2}}
\end{equation}
We first note that the Laplacian including self loops is the same as the regular Laplacian:
\begin{equation}
    \label{appendixEq:laplacian}
    \tilde{\bfL} = \tilde{\bfD} - \tilde{\bfA} = \bfD + \bfI - \bfA - \bfI  = \bfD - \bfA = \bfL.
\end{equation}
Therefore, it holds that:
\begin{eqnarray}
    \label{appendixEq:proof}
    \nonumber
    \tilde \bfP &=& \bfI - \bfI +  \tilde{\bfD}^{-\frac{1}{2}} \tilde{\bfA} \tilde{\bfD}^{-\frac{1}{2}} \nonumber\\ 
    &=&  \bfI -  \tilde{\bfD}^{-\frac{1}{2}} \tilde{\bfD} \tilde{\bfD}^{-\frac{1}{2}} +  \tilde{\bfD}^{-\frac{1}{2}} \tilde{\bfA} \tilde{\bfD}^{-\frac{1}{2}} \nonumber \\
    &=&   \bfI - \tilde{\bfD}^{-\frac{1}{2}} (\tilde{\bfD} - \tilde{\bfA}) \tilde{\bfD}^{-\frac{1}{2}} \\
    &=& \bfI - \tilde{\bfD}^{-\frac{1}{2}} (\bfD - \bfA) \tilde{\bfD}^{-\frac{1}{2}} \nonumber \\
    &=& \bfI - \tilde{\bfD}^{-\frac{1}{2}} \bfL \tilde{\bfD}^{-\frac{1}{2}}. \nonumber
\end{eqnarray}

\subsection*{Proof of Theorem 2.1}
\begin{proof}
First, note that  \eqref{eq:dirichletEnergy} from the main paper can be written as
\begin{equation}
    \label{appendixeq:dirichletEnergy}
    E(\bff^{(l)}) = \sum_{i \in \mathcal{V}} \sum_{j \in \mathcal{N}_i} \frac{1}{2} \textstyle{\left\|\frac{\bff^{(l)}_i}{\sqrt{(1+d_i)}} - \frac{\bff^{(l)}_j}{\sqrt{(1+d_j)}} \right\|_2^2} = \frac{1}{2}\|\bfG\tilde\bfD^{-\frac{1}{2}}\bff^{(l)}\|_2^2,
\end{equation}
where $\bfG$ is the graph gradient operator, also known as the incidence matrix, that for each edge subtracts the features of the two connected nodes, i.e., $\bfG \bff^{(l)}_{(i,j)} = \bff^{(l)}_i - \bff^{(l)}_j$ for $(i,j)\in\mathcal{E}$. Let us assume that the initial feature $\bff^{(0)}$ has some Dirichlet energy $E_0 > E_{opt}$ as defined in  \eqref{appendixeq:dirichletEnergy}. Since 
$$
\nabla E = \tilde\bfD^{-\frac{1}{2}}\bfG^\top\bfG\tilde\bfD^{-\frac{1}{2}}\bff^{(l)}
$$
we see that the forward propagation through a GCN approximates the gradient flow of the Dirichlet energy. That is, for given $L$ and and $\omega$ we have that 
\begin{equation}
\label{eq:SD_E}
\bff^{(l+1)}= \bff^{(l)} - \omega\nabla E = \bff^{(l)} - \omega\tilde\bfD^{-\frac{1}{2}}\bfG^\top\bfG\tilde\bfD^{-\frac{1}{2}}\bff^{(l)} = (\bfI - \omega \tilde\bfD^{-\frac{1}{2}} \bfL\tilde\bfD^{-\frac{1}{2}})\bff^{(l)}
\end{equation}
where we used that $\bfG^{\top}\bfG = \bfL$.
 Equation \ref{eq:SD_E} can be seen both as a gradient descent step to reduce $E$, and also as a forward Euler approximation with step size $\omega$ of the solution of 
\begin{equation}
\label{eq:ODE}
\frac{\partial \bff(t)}{\partial t} = -\tilde\bfD^{-\frac{1}{2}} \bfL\tilde\bfD^{-\frac{1}{2}}\bff(t), \quad \bff(0) = \bff^{(0)}.
\end{equation}

It is known that the solution to \eqref{eq:ODE} is given by %\cite{evans10}
\begin{equation}\label{eq:mat_exp}
\bff(t) = \exp\left(-t\tilde\bfD^{-\frac{1}{2}} \bfL\tilde\bfD^{-\frac{1}{2}}\right)\bff(0).
\end{equation}
Since the Dirichlet energy of $\bff(t)$ is continuous in $t$ and decays monotonically from $E_0$ to zero, there exists a $T$ such that $E(\bff (T))$ = $E^*$.
%Assume that the optimal energy is achieved at the time $T$, i.e., $E_{opt} = E(\bff(T))$. 
Now, considering discrete time intervals $0=t_0,...,t_L=T$, then, similarly to  \eqref{eq:mat_exp}, for any two subsequent time steps $t_{l+1}$ and $t_l$ we have that
\begin{equation}
\bff(t_{l+1}) = \exp\left(-(t_{l+1}-t_l)\tilde\bfD^{-\frac{1}{2}} \bfL\tilde\bfD^{-\frac{1}{2}}\right)\bff(t_l).
\end{equation}
Taking fixed-interval time steps such that $t_{l+1} - t_{l} = \omega = T/L$ for $l=0,...,L$, we get 
\begin{equation}\label{eq:EulerTaylor}
\bff(t_{l+1}) = \exp\left(-\omega\tilde\bfD^{-\frac{1}{2}} \bfL\tilde\bfD^{-\frac{1}{2}}\right)\bff(t_{l})= (\bfI - \omega \tilde\bfD^{-\frac{1}{2}} \bfL\tilde\bfD^{-\frac{1}{2}})\bff(t_l) + O(\omega^2),
\end{equation}
where the rightmost approximation holds due to the Taylor expansion, up to first-order approximation.
Denoting $\bff^{(l)} = \bff(t_l)$ and $\bar\omega = T$, we complete the proof. \end{proof}

\begin{remark}
\tred{
At the basis of our analysis above there is the analytical solution in \eqref{eq:mat_exp}, which, as shown in Eq. \eqref{eq:EulerTaylor}, is $O(\omega^2)$ different than the propagated solution through \eqref{eq:SD_E}. After $L$ layers, the $O(\omega^2)$ term may accumulate $L$ times. Since $\omega=T/L$ where $T$ is fixed, then $O(L\omega^2)$ is equivalent to $O(\omega)$, resulting in a first order approximation to the analytical solution in \eqref{eq:mat_exp}. This is often referred to as forward Euler integration. To demonstrate and verify this, we perform a small experiment with a random and diagonally normalized symmetric positive definite matrix $\bfA$ (in the role of symmetric normalized Laplacian). See Fig.  \ref{fig:EulerVerify} for details. Indeed, the difference between the analytical and propagated solutions at the last layer scales as $\omega$ or $1/L$ where $L$ is the number of layers. }
\end{remark}

\begin{figure}[t]
\centering
\begin{tikzpicture}
  \begin{axis}[
      width=0.5\linewidth, 
      height=0.35\linewidth,
      grid=major,
      ymode=log,
      xmode=log,
      grid style={dashed,gray!30},
      xlabel=Layers,
      %ylabel= $\|\bff(T)-\bff^{(L)}\|_2$,
      ylabel near ticks,
      legend style={at={(0.77,0.999)},anchor=north,scale=1.0, draw=none, cells={anchor=west}, font=\small, fill=none},
      legend columns=1,
      xtick={4,8,16,32,64,128,256},
      xticklabels = {4,8,16,32,64,128,256},
      yticklabel style={
        /pgf/number format/fixed,
        /pgf/number format/precision=3
      },
      scaled y ticks=false,
      every axis plot post/.style={thick},
      ymax=1.0,
      ymin=0,
      xmin = 3,
      xmax = 280,
    ]
    \addplot[red, mark=oplus*, forget plot]
    table[x=layer,y=Difference,col sep=comma] {data/EulerVerification.csv};
    \addplot[blue, style=dotted , forget plot]
    table[x=layer,y=1/L,col sep=comma] {data/EulerVerification.csv};

    %\addplot[magenta, style=dotted, mark=oplus*, forget plot]
    %table[x=layer,y=ref,col sep=comma] {data/omega_sum.csv};
    
    \addplot[red, draw=none] coordinates {(1,1)};
    \addplot[blue, style=dotted, draw=none] coordinates {(1,1)};
     
    \legend{$\|\bff(T)-\bff^{(L)}\|_2$, $1/L$}

    \end{axis}
\end{tikzpicture}
\caption{\tred{Difference norm $\|\bff(T)-\bff^{(L)}\|_2$ between an analytical solution $\bff(t) = \exp\left(-t \bfA\right)\bff(0)$ like in \eqref{eq:mat_exp} and a propagated solution $\bff^{(l+1)}=(\bfI-\omega\bfA)\bff^{(l)}$ like in \eqref{eq:SD_E} using a random and diagonally normalized symmetric positive definite matrix $\bfA\in\mathbb{R}^{100\times 100}$, and a random initial feature $\bff(0)$. The integration goes from $t=0$ to $T=1$, and hence $\omega = 1/L$. It is clear that the difference norm at the last layer scales like $1/L$, as expected.}}
\label{fig:EulerVerify}
\end{figure}

%\begin{corollary}
%\label{cor:sumOmega}
%Allowing a variable $\omega^{(l)} > 0$ at each layer in Theorem \ref{theorem:single_omega}, yields $\sum_{l=0}^{L-1} \omega^{(l)} = \bar\omega$ up to first order accuracy. 
%\end{corollary}

%\begin{cor}
%\tred{COPY LATER FROM TEXT}
%Allowing a variable $\omega^{(l)} > 0$ at each layer in Theorem \ref{theorem:single_omega}, yields that $\sum_{l=0}^{L-1} \omega^{(l)}$ \tred{converges to a constant independent of $L$ up to first order accuracy}. 
%\end{cor}
\subsection*{Proof of Corollary 2.2}
\begin{proof}
The proof follows immediately by setting variable $t_{l+1} - t_{l} = \omega^{(l)}$ and placing in  \eqref{eq:EulerTaylor}. 
\end{proof}

\begin{remark}[The non-negativity of $\tilde\bfP_\omega$]
By definition, for $0 < \omega \leq 1$ all the spatial weights of     $\tilde\bfP_\omega$ defined in  \eqref{eq:P_omega} are non-negative, and it is that the operator is smoothing as it is a low-pass filter. For $\omega > 1$ or $\omega < 0$, by definition we have an operator with mixed signs. 
\end{remark}

%\begin{thm}
%Assume that the graph is connected. Then, there exists some $\omega_0\geq1$ where for all $0<\omega<\omega_0$, the operator $\tilde\bfP_\omega$ in  \eqref{eq:P_omega} from the main paper is smoothing and the leading eigenvector is  $\tilde\bfD^{\frac{1}{2}}\mathbf{1}$. For $\omega > \omega_0$ or $\omega < 0$, the leading eigenvector changes.
%\end{thm}
\subsection*{Proof of Theorem 2.3}
\begin{proof}
Assuming that the graph is connected, it is known that the graph Laplacian matrix has the eigenvector $\mathbf{1}$ whose eigenvalue is $0$, i.e. $\bfL\mathbf{1}=0$. Hence, we get that $\tilde\bfD^{-\frac{1}{2}} \bfL\tilde\bfD^{-\frac{1}{2}}\tilde\bfD^{\frac{1}{2}}\mathbf{1} = 0$ so $\tilde{\bfD}^{\frac{1}{2}}\mathbf{1}$ is the eigenvector of the normalized Laplacian with eigenvalue of 0. 

Furthermore, denote the normalized Laplacian by $\tilde{\bfL} = \tilde\bfD^{-\frac{1}{2}} \bfL\tilde\bfD^{-\frac{1}{2}}$. Consider the range
$$ 
0<\omega<\frac{2}{\rho(\tilde\bfL)} = \omega_0,
$$
where $\rho(\tilde\bfL)$ denotes the spectral radius of the matrix $\tilde\bfL$.
It is easy to verify that for this range of values for $\omega$, the largest eigenvalue in magnitude of $\tilde\bfP_\omega$ is 1, and it corresponds to the null eigenvector of $\tilde\bfL$, i.e., $\tilde{\bfD}^{\frac{1}{2}}\mathbf{1}$. Hence, for this range, $\tilde\bfP$ is smoothing. 
%hence in this range $\omega\leq 1$, and by definition, $\tilde\bfP_\omega$ is a non-negative matrix. 
For $\omega > \omega_0$ and $\omega < 0$, the leading eigenvector of $\tilde\bfP_\omega$ becomes the leading eigenvector of $\tilde\bfL$.
Furthermore, it can be shown that $\rho(\tilde\bfL)\leq 2$ (see \cite{Williamson2016}
 for the proof), hence $\omega_0 \geq 1$.
\end{proof}

\section{Synthetic expressiveness task}
\label{appendix:synthetic_task}
To demonstrate the importance and benefit of learning sharpening propagation operators in addition to smoothing operators, we propose the following synthetic node gradient regression task. Given a graph $\mathcal{G}=(\mathcal{V}, \mathcal{E})$ with some input node features $\bff^{in} \in \mathbb{R}^{n \times c_{in}}$, we wish a GNN to regress the node features gradient, $\nabla \bff^{in}$, where the node feature gradient of the $i$-th node is defined as an upwind gradient operator:
\begin{equation}
    \label{eq:nodeGrad}
    \nabla \bff^{in}_i = \rm{max}_{j \in \mathcal{N}_i}  (\bff_i - \bff_j) ,
\end{equation}
where the goal of the considered GNN is to minimize the following objective:
\begin{equation}
    \label{eq:syntheticgoal}
    \| \rm{GNN}(\bff^{in}, \mathcal{G}) - \nabla\bff^{in}\|_2^2. 
\end{equation}

As a comparison, we consider two GNNs: GCN \citep{kipf2016semi} and our $\omega$GCN with 64 channels and 2 layers. In both cases we use a learning rate of $1e-4$ without weigh decay and train the network for 5000 iterations (no further benefit was obtained with any of the considered methods). The input graph is a random Erdős–Rényi graph with 8 nodes and an edge rate of $30\%$, with input node features sampled form a uniform distribution in the range of 0 to 1.  The obtained loss of GCN is of order  $1e-1$, while our $\omega$GCN obtains a loss of order $1e-12$, also as can be seen in Fig. \ref{fig:gradEstimationExperiment}. We therefore conclude that introducing the ability of learning mixed-sign operators by $\omega$ is beneficial to enhance the expressiveness of GNNs.

\section{Datasets}
\label{appendix:datasets}
In this section we provide the statistics of the datasets used throughout our experiments. Tab. \ref{table:nodeclassification_datasets} presents information regarding node-classification datasets, and Tab. \ref{table:graphclassification_datasets} summarizes the graph-classification datasets. For each dataset, we also provide the homophily score as defined by \cite{Pei2020Geom-GCN:}.

\begin{table}[h]
  \caption{Node classification datasets statistics. Hom. score denotes the homophily score.}
  \label{table:nodeclassification_datasets}
  \begin{center}
  \resizebox{1.0\linewidth}{!}{\begin{tabular}{lcccccccccc}
  \toprule
    Dataset & Cora & Citeseer & Pubmed & Chameleon & Film  & Cornell & Texas & Wisconsin & PPI & Ogbn-arxiv  \\
    \midrule
    Classes & 7 & 6 & 3 & 5 & 5 & 5 & 5 & 5 & 121 & 40 \\
    Nodes & 2,708 & 3,327  & 19,717 & 2,277 & 7,600 & 183 & 183 & 251 & 56,944 & 169,343   \\
    Edges & 5,429 & 4,732 & 44,338 & 36,101 & 33,544 & 295 & 309 & 499 & 818,716 & 1,116,243  \\
    Features & 1,433 & 3,703 &  500 & 2,325 &  932 &   1,703 & 1,703 & 1,703 & 50 & 128\\
    Hom. score & 0.81 & 0.80 & 0.74 & 0.23 & 0.22 &  0.30 & 0.11 & 0.21 & 0.17  &  0.63 \\
    \bottomrule
  \end{tabular}}
  \end{center}
\end{table}

\begin{table}[h]
  \caption{TUDatasets graph classification statistics.}
  \label{table:graphclassification_datasets}
  \begin{center}
  \begin{tabular}{lcccccc}
  \toprule
    Dataset & MUTAG & PTC & PROTEINS & NCI1 & NCI109  \\
    \midrule
    Classes & 2 & 2 & 2 & 2 & 2\\
    Graphs & 188 & 344 & 1113 & 4110 & 4127\\
    Avg. nodes & 17.93 & 14.29 & 39.06 & 29.87 & 32.13 \\
    Avg. edges & 19.79 & 14.69 & 72.82 & 32.30 & 32.13 \\
    \bottomrule
  \end{tabular}
  \end{center}
\end{table}

\section{Over-smoothing in GAT}
\label{appendix:gat_oversmoothing}
In addition to the observation presented in Sec. \ref{sec:wgnns_details} and specifically in Fig. \ref{fig:energyFig1} where we see that recurrent applications of GAT reduces the node feature energy from  \eqref{eq:gatEnergy}, which causes over-smoothing as shown by \cite{wu2019simplifying, wang2019improving} (as discussed in the main paper), here, we also show that the same behaviour is evident with Citeseer and Pubmed datasets in Fig. \ref{fig:appendixGATOVERSMOOTHING}.

\input{images/FigEnergyCiteSeer_PubMed.tex}

\section{Architectures in details}
\label{appendix:architectures}
We now elaborate on the specific architectures used in our experiments in Sec. \ref{sec:experiments}. As noted in the main paper, all our network architectures consist of an opening (embedding) layer ($1
\times 1$ convolution), a sequence of $\omega$GNN (i.e., $\omega$GCN or $\omega$GAT) layers, and a closing (classifier) layer ($1
\times 1$ convolution).
In total, we have two types of architectures -- one that is based on GCN, for node classification tasks reported in Tab. \ref{table:nodeClassificationArch}, and the other for the graph classification task which is based on \cite{xu2018how} and is reported in Tab. \ref{table:graphClassificationArch}. Throughout the following, we denote by $c_{in}$ and $c_{out}$ the input and output channels, respectively, and $c$ denotes the number of features in hidden layers (which is a reported in Appendix \ref{appendix:hyperparameters}). We initialize the embedding and classifier layers with the Glorot \citep{glorot2010understanding}
initialization, and $\bfK^{(l)}$ from  \eqref{eq:omegaGNN} is initialized with an identity matrix of shape $c \times c$. The initialization of $\Omega^{(l)}$ also starts from a vectors of ones. We note that our initialization yields a standard smoothing process, which is then adapted to the data as the learning process progresses, and if needed also changes the process to a non-smoothing one by the means of mixed-signs, as discussed earlier and specifically in Theorem. \ref{theorem:omega_smoothing}. We denote the number of $\omega$GNN layers by $L$, and the dropout probability by $p$. The main difference between the two architectures are as follows. First, for the graph classification we use the standard add-pool operation as in GIN \citep{xu2018how} to obtain a global graph feature. Second, we follow GIN and in addition to the graph layer (which is $\omega$GNN in our work), we add batch normalization (denoted by BN), $1\times1$ convolution and a ReLU activation past each graph layer.

\begin{table}[h]
  \caption{The architecture used for node classification and inductive learning.}
  \label{table:nodeClassificationArch}
  \begin{center}
  \begin{tabular}{lcc}
  \toprule
    Input size & Layer  &  Output size \\
    \midrule
    $n \times c_{in}$ & $1\times1$ Dropout(p) & $n \times c_{in}$ \\
    $n \times c_{in}$ & $1\times1$ Convolution & $n \times c$ \\
    $n \times c$ & ReLU & $n \times c$ \\    
    $n \times c$ & $L \times $ $\omega$GNN layers & $n \times c$ \\
    $n \times c$ &  Dropout(p) & $n \times c$ \\
    $n \times c$ & $1\times1$ Convolution & $n \times c_{out}$ \\
    \bottomrule
  \end{tabular}
\end{center}
\end{table}

\begin{table}[h]
  \caption{The architecture used for graph classification.}
  \label{table:graphClassificationArch}
  \begin{center}
  \begin{tabular}{lcc}
  \toprule
    Input size & Layer  &  Output size \\
    \midrule
    $n \times c_{in}$ & $1\times1$ Convolution & $n \times c$ \\
    $n \times c$ & ReLU & $n \times c$ \\    
    $n \times c$ & $L  \times [ $ $\omega$GNN , BN,  $1\times1$ Convolution, ReLU $]$  & $n \times c$  \\
    $n \times c$ & $1\times1$ Add-pool & $1 \times c$ \\
    $1 \times c$ & $1\times1$ Convolution & $1 \times c$ \\

    $1 \times c$ & $1\times1$ Dropout(p) & $1 \times c$ \\
    $1 \times c$ & $1\times1$ Convolution & $1 \times c_{out}$ \\
    \bottomrule
  \end{tabular}
\end{center}
\end{table}

\section{Hyper-parameters details}
\label{appendix:hyperparameters}
We provide the selected hyper-parameters in our experiments. We denote the learning rate of our $\omega$GNN layers by $LR_{GNN}$, and the learning rate of the $1\times 1$ opening and closing as well as any additional classifier layers by $LR_{oc}$.
Also, the weight decay for the opening and closing layers is denoted by $WD_{oc}$.  We denote the $\omega$ parameter learning rate and weight decay by $LR_{\omega}$ and $WD_{\omega}$, respectively. $c$ denotes the number of hidden channels. In the case of $\omega$GAT, the attention head vector $\bfa$ are learnt with the same learning rate as $LR_{GNN}$ and $WD_{GNN}$. 

\subsection{Semi-supervised node classification}

The hyper-parameters for this experiment are summarized in Tab. \ref{table:semisupervisedHyperParams}.

\begin{table}[ht!]
  \caption{Semi-supervised node classification hyper-parameters.}
  \label{table:semisupervisedHyperParams}
  \begin{center}
  %\resizebox{1.0\linewidth}{!}{
  \begin{tabular}{llccccccccc}
  \toprule
    Architecture & Dataset & $LR_{GNN}$ & $LR_{oc}$ & $LR_{\omega}$ & $WD_{GNN}$ & $WD_{oc}$ &  $WD_{\omega}$ &  $c$ & $p$  \\
    \midrule
    $\omega$GCN & Cora & 0.01 & 0.01 & 0.01 & 1e-4 & 8e-5 & 2e-4 & 64 & 0.6 \\
    &Citeseer & 1e-4 & 0.005  & 0.005 & 1e-5 &  5e-6 & 2e-4 & 256 & 0.7    \\
    & Pubmed &  0.001 & 5e-4 & 0.005 & 2e-4 & 1e-4 & 1e-4 & 256 & 0.5  \\
    \midrule   
    $\omega$GAT  & Cora & 0.01 & 0.01 & 0.005 & 1e-5 & 1e-5 & 1e-5 &  64 & 0.6 \\
    &Citeseer & 0.005 & 0.005 & 0.001 & 1e-4 & 1e-5 & 1e-4 & 256 & 0.7   \\
    & Pubmed & 0.005 & 0.001 & 0.05 & 4e-5 & 1e-5 & 1e-4 & 256 & 0.5 \\
    \bottomrule
  \end{tabular}
  %}
\end{center}
\end{table}

\subsection{Full-supervised node classification}
The hyper-parameters for this experiment are summarized in Tab. \ref{table:fullysupervisedHyperParameters}. For Ogbn-arxiv from Tab. \ref{table:arxiv}, 8 layer $\omega$GCN and $\omega$GAT were employed

\subsection{Inductive learning} The hyper-parameters for the inductive learing on PPI are listed in Sec. \ref{sec:inductivelearningppi} in the main paper, and are the same for $\omega$GCN and $\omega$GAT.

\begin{table}[ht!]
  \caption{Full-supervised node classification hyper-parameters.}
\label{table:fullysupervisedHyperParameters}
  \begin{center}
  %\resizebox{1.0\linewidth}{!}{
  \begin{tabular}{llccccccccc}
  \toprule
   Architecture &  Dataset & $LR_{GNN}$ & $LR_{oc}$ & $LR_{\omega}$ & $WD_{GNN}$ & $WD_{oc}$ &  $WD_{\omega}$ & $c$ & $p$  \\
    \midrule
    $\omega$GCN & Cora & 0.01 & 0.05 & 0.005 & 0.01 & 1e-4 & 1e-4 & 64 & 0.5   \\
    &Citeseer & 0.001 & 0.08 & 0.005 & 0.005 & 1e-4 & 0 & 64 & 0.5   \\
    &Pubmed & 0.005 & 0.005 & 0.01 & 0.003 & 5e-5 & 0.01 & 64 & 0.5 \\
    &Chameleon & 1e-4 & 0.005 & 5e-4 & 1e-4 & 1e-4 & 1e-5 & 64 & 0.5 \\
    &Film & 0.05 & 0.01 & 0.05 & 1e-4 & 1e-4 & 1e-5 & 64 & 0.5 \\
    &Cornell & 0.01 & 0.05 & 0.01 & 0.005 & 1e-4 & 0 & 64 & 0.5  \\
    &Texas & 0.08 & 0.08 & 0.005 &  0.005 & 5e-4 & 0 & 64 & 0.5  \\
    &Wisconsin & 0.001 & 0.05 & 0.005 & 1e-4 & 3e-4 & 3e-4 & 64  & 0.5  \\
    &Ogbn-arxiv & 0.01 & 0.01 & 0.01 & 0 & 0 & 0 & 256 & 0 \\
    \midrule
        $\omega$GAT &Cora & 0.001  & 0.01 & 0.05 & 0.001 & 5e-4 & 0 & 64 & 0.5  \\
    &Citeseer & 0.005 & 0.05 & 0.03 & 0.005 & 5e-4 & 0.001 & 64 & 0.5 \\
    &Pubmed & 0.05 & 0.005 & 0.005 &  0.003 & 1e-6 & 0.003 & 64 & 0.5 \\
    &Chameleon & 0.005 & 0.005 & 3e-4 & 5e-4 &  5e-4 & 1e-5   \\
    &Film & 0.05 & 0.01 & 0.01 & 5e-4 & 0.001 & 4e-4 & 64 & 0.5  \\
    &Cornell & 0.001 & 0.01 & 0.005 & 1e-4 & 1e-5 & 0 & 64 & 0.5\\
    &Texas & 1e-4 & 0.02 & 0.05 & 5e-4 & 5e-4 & 0 & 64 & 0.5 \\
    &Wisconsin & 0.01 & 0.05 & 0.005 & 0.001 & 5e-4 &  0 & 64 & 0.5 \\
    &Ogbn-arxiv & 0.01 & 0.01 & 0.01 & 0 & 0 & 0 & 256 & 0 \\
    \bottomrule
  \end{tabular}
  %}
\end{center}
\end{table}

\subsection{Graph classification}
The hyper-parameters for the graph classification experiment on TUDatasets are reported in Tab. \ref{table:graphClassificationHyperParams}. We followed the same grid-search procedure as in GIN \citep{xu2018how}. In all experiment, a 5 layer (including the initial embedding layer) $\omega$GCN and $\omega$GAT are used, similarly to GIN.

\begin{table}[ht!]
  \caption{Graph classification hyper-parameters. BS denoted batch size.}
  \label{table:graphClassificationHyperParams}
  \begin{center}
  %\resizebox{1.0\linewidth}{!}{
  \begin{tabular}{llcccccccccc}
  \toprule
    Architecture & Dataset & $LR_{GNN}$ & $LR_{oc}$ & $LR_{\omega}$ &  $WD_{GNN}$  &  $WD_{oc}$ &  $WD_{\omega}$ &   $c$ & $p$ & BS \\
    \midrule
  $\omega$GCN &  MUTAG & 0.01 & 0.01 & 0.01 & 0 & 0 & 0  & 32 & 0 & 32 \\
    &PTC &  0.01 & 0.01 & 0.01 & 0 & 0  & 0 & 32 & 0 & 32   \\
    &PROTEINS &  0.01 & 0.01 & 0.01 &  0 & 0 & 0 & 32 & 0 & 128 \\
    &NCI1 &  0.01 & 0.01 & 0.01 & 0 &  0 & 0 & 32 & 0.5 & 32\\
    &NCI109 &  0.01 & 0.01 & 0.01 & 0 & 0 & 0 & 32 & 0 & 32 \\
    \midrule
      $\omega$GAT &  MUTAG &  0.01 & 0.01 & 0.01 & 0 & 0 & 0  & 32 & 0 & 32 \\
    &PTC &  0.01 & 0.01 & 0.01 & 0 & 0 & 0  & 32 & 0 & 128  \\
    &PROTEINS &   0.01 & 0.01 & 0.01 & 0  & 0 & 0 & 32 & 0 & 128\\
    &NCI1 &  0.01 & 0.01 & 0.01 & 0 & 0  & 0 & 32 & 0.5 & 128\\
    &NCI109 &  0.01 & 0.01 & 0.01 & 0  & 0 & 0 & 32 & 0.5 & 32 \\
    \bottomrule
  \end{tabular}
  %}
\end{center}
\end{table}

\subsection{Ablation study}
In this experiment we used the same hyper-parameters as reported in Tab. \ref{table:semisupervisedHyperParams}.

\section{Runtimes}
\label{appendix:runtimes}

Following the computational cost discussion from Sec. \ref{sec:computationalcost} in the main paper, we also present in Tab. \ref{tab:runtimes} the measured training and inference times of our baselines GCN and GAT with 2 layers, where we see that indeed the addition of $\omega$ per layer and channel requires a negligible addition of time, at the return of a significantly more accurate GNN. We note that further accuracy gain can be achieved when adding more $\omega$GNN layers as reported in Tab. \ref{table:semisupervised} in the main paper. However, since GCN and GAT over-smooth, the comparison here is done with 2 layers, where the highest accuracy is obtained for the baseline models.  %Also, we compare our runtime with two other popular methods - GCNII and PDE-GCN. Specifically, we see that PDE-GCN requires a significant amount of computation time - due to its convolution in \emph{edge} space as opposed to the typical \emph{vertex} space (as in GCN, GAT, GCNII and our $\omega$GNNs). 
%\tred{The total runtime for all the experiments conducted in this paper, including all hyper-parameters grid search required a 50 days worth of computation of an Nvidia Titan RTX with 24GB of memory.}

\begin{table}[h]
  \caption{Training and inference GPU runtimes [ms] on Cora.}
  \label{tab:runtimes}
  \begin{center}
  \begin{tabular}{lcccccc}
  \toprule
   Runtime  & GCN & GAT   & $\omega$GCN (Ours) & $\omega$GAT (Ours)  \\
    \midrule
    Training & 7.71 & 14.59  & 7.79 & 14.95  \\
    Inference  & 1.75 & 2.98   &  1.88 & 3.09 \\
    Accuracy ($\%$) &  81.1 & 83.1  & 82.6 & 83.4 \\
    \bottomrule
  \end{tabular}
  \end{center}
\end{table}

\section{Ablation study using $\omega$GAT}
\label{appendix:ablation}
To complement our ablation study on $\omega$GCN in Sec. \ref{sec:ablation} in the main paper, we perform as similar study on $\omega$GAT. Here, we show in Tab. \ref{tab:omegaGAT_ablation}, that indeed the single $\omega$ variant, dubbed $\omega$GAT\textsubscript{G} does not over-smooth, and that by allowing the greater flexibility of a per-layer and per layer and channel of our $\omega$GAT\textsubscript{PL} and $\omega$GAT, respectively, better performance is obtained.

\begin{table}[ht!]
  \caption{Ablation study on $\omega$GAT.}
  \label{tab:omegaGAT_ablation}
\begin{center}
  
  \setlength{\tabcolsep}{2.mm}{\begin{tabular}{llcccccc}
    \toprule
    \multirow{2}{*}{Data.} & \multirow{2}{*}{Variant} & \multicolumn{6}{c}{Layers} \\
                         &  & 2  & 4  & 8 & 16 & 32 & 64 \\
    \midrule
    Cora & $\omega$GAT\textsubscript{G} &  83.3 & 83.3 & 83.4 & 83.6 & 83.7 &83.9
     \\
    & $\omega$GAT\textsubscript{PL} & 83.4& 83.5 & 83.8 & 84.0 & 84.1&  84.0
    \\
    & $\omega$GAT & 83.4 & 83.7 & 84.0 & 84.3 & 84.4 & 84.8  \\
   \midrule
    Cite. & $\omega$GAT\textsubscript{G} & 71.5& 71.8& 71.9& 72.2& 72.4& 72.9

   \\
    & $\omega$GAT\textsubscript{PL} & 72.1& 72.3& 72.4&	72.8& 73.1&	73.2
  \\
    & $\omega$GAT &  72.5 & 73.1 & 73.3 & 73.5 & 73.9 & 74.0
    \\
    \midrule
        Pub. & $\omega$GAT\textsubscript{G} & 80.0&	80.2& 80.3& 80.5& 80.6& 80.9
  \\
    & $\omega$GAT\textsubscript{PL} & 80.0& 80.4& 80.7& 81.1& 81.2& 81.4
  \\
    & $\omega$GAT & 80.3 & 81.0 &  81.2 & 81.3 & 81.5 & 81.8 \\

    \bottomrule
  \end{tabular}}
  \end{center}
\end{table}

\section{Statistical significance of semi-supervised node classification results}
\label{appendix:StatisticalSig}
\begin{table}[h]
  \caption{Semi-supervised node classification test accuracy 100 random train-val-test splits. }
  \label{tab:significance}
  \begin{center}
  \begin{tabular}{lccc}
    \toprule
    Method & Cora & Citeseer & Pubmed \\
        \midrule
    GCN \citep{kipf2016semi} & 81.5& 71.9 & 77.8 \\
    GAT \citep{velickovic2018graph} & 81.8 & 71.4 & 78.7\\
    MoNet \citep{monti2017geometric} & 81.3 &71.2 & 78.6\\
    GRAND-l \citep{chamberlain2021grand} & 83.6 & 73.4 & 78.8 \\
    GRAND-nl \citep{chamberlain2021grand} & 82.3 & 70.9 & 77.5\\
    GRAND-nl-rw \citep{chamberlain2021grand} & 83.3 & 74.1 & 78.1 \\
    GraphCON-GCN \citep{rusch2022graph} & 81.9 & 72.9 & 78.8 \\
    GraphCON-GAT \citep{rusch2022graph} & 83.2 & 73.2 & 79.5 \\
    GraphCON-Tran \citep{rusch2022graph} & 84.2 & \textbf{74.2} & 79.4 \\
    \midrule
    $\omega$GCN (ours) & \textbf{84.5} & 73.8  & \textbf{82.9} \\
    $\omega$GAT (ours) & 84.3 & 73.6 & 82.6 \\
    \bottomrule
  \end{tabular}
\end{center}
\end{table}

Throughout our semi-supervised node classification experiment in Sec. \ref{sec:experiments} on Cora, Citeseer and Pubmed, the standard split from \cite{kipf2016semi} was considered, to a direct comparison with as many as possible methods. However, since this result reflects the accuracy from a single split, we also repeat this experiment with 100 random splits as in \cite{chamberlain2021grand} and compare with applicable methods that also conducted such statistical significance test. In Tab. \ref{tab:significance}, we report our obtained accuracy on Cora, Citeseer and Pubmed. It is possible to see that in this experiment our $\omega$GCN and $\omega$GAT outperform or obtain similar results compared with the considered methods, which further highlight the performance advantage of our method.

\begin{table}[t]
  %\parbox{0.48\linewidth}{\caption{Fully-supervised node classification accuracy ($ \%$).}
  \caption{Node classification accuracy (\%) on additional datasets.}
  \centering
  \label{table:arxiv}
  \begin{tabular}{lc}
    \toprule
    Method & Ogbn-arxiv \\
        \midrule
    GCN \citep{kipf2016semi}  &   71.74  \\
    GAT \citep{velickovic2018graph}  & 71.59   \\
    GATv2 \citep{brody2022how_GATV2}  & 71.87   \\
    APPNP \citep{klicpera2018combining} & 71.82 \\
    Geom-GCN-P \citep{Pei2020Geom-GCN:}  & -- \\
    JKNet \citep{jknet}    & 72.19 \\
    SGC \citep{wu2019simplifying}   & 69.20 \\
    GCNII \citep{chen20simple}   & 72.74 \\
    EGNN \citep{zhou2021dirichlet}  & 72.70  \\ 
    GRAND \citep{chamberlain2021grand}  & 72.23 \\
    AGDN \cite{zhao2021adaptive} & \textbf{73.41} \\
    \midrule
    $\omega$GCN (Ours)   &\ 73.02   \\
    $\omega$GAT (Ours)   & 72.76 \\
    \bottomrule
  \end{tabular}
 \end{table}

\section{The learnt $\vec{\omega}$}
\label{appendix:learntOmega}
\begin{figure}[h]
    \centering
    \includegraphics[width=0.6\textwidth]{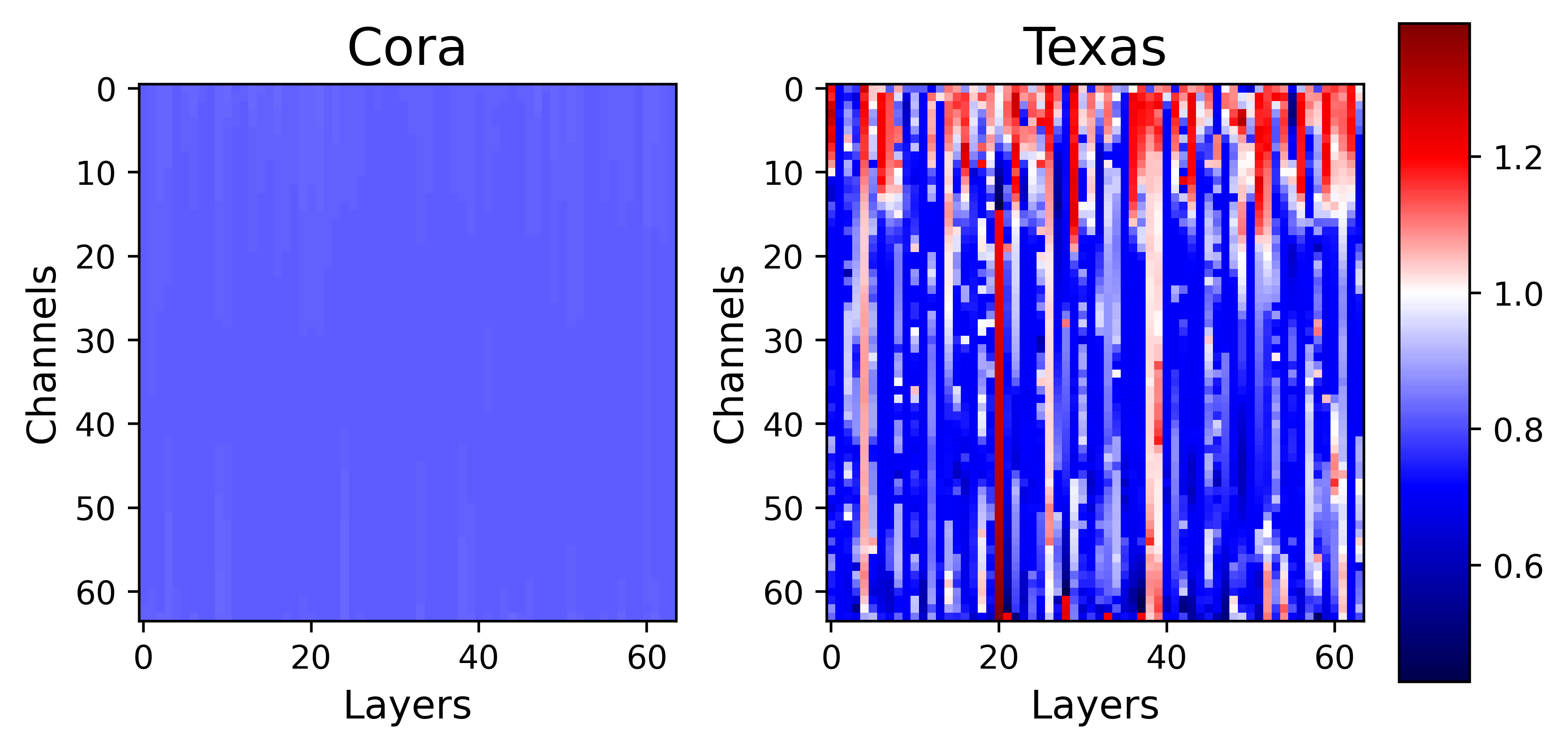}
    \caption{The learnt $\vec{\omega} \in \mathbb{R}^{64\times64}$ of $\omega$GCN with 64 layers (x-axis) and 64 channels (y-axis) for Cora (homophilic) and Texas (heterophilic) datasets. Smoothing operators appear in blue, while sharpening operators appear in red. White entries are obtained for $\omega=1$.}
\label{fig:learntOmegaKernels}
\end{figure}

One of the main advantages of our method in Sec. \ref{sec:wgnns} is that our method is capable of learning both smoothing and sharpening propagation operators, which cannot be obtained in most current GNNs. In Fig. \ref{fig:learntOmegaKernels} we present the actual $\{\vec{\omega^{(l)}}\}^L_{l=1}$ as a  matrix of size $L \times c$ that was learnt for two dataset of different types---with high and low homophily score (as described in \cite{Pei2020Geom-GCN:}). Namely, the Cora dataset with a high homophily score of 0.81, and the Texas dataset with a low homophily score of 0.11 (i.e., a heterophilic dataset). We see that a homophilic dataset like Cora, the network learnt to perform diffusion, albeit in a controlled manner, and not to simply employ the standard averaging operator $\tilde{\bfP}$. We can further see that for a heterophilic dataset the ability to learn contrastive (i.e., sharpening) propagation operators in addition to diffusive kernels is beneficial, and is also reflected in our results in Tab. \ref{table:homophilic_fully}-\ref{table:heterophilic_fully}, where a larger improvement is achieved in datasets like Cornell, Texas and Wisconsin, which have low homophily scores \citep{rusch2022graph}.

\end{document}

%% file: images/FigEnergy.tex
\begin{figure*}
\centering
\begin{minipage}{0.48\textwidth}
\centering
\begin{tikzpicture}[define rgb/.code={\definecolor{mycolor}{RGB}{#1}},
                    rgb color/.style={define rgb={#1},mycolor}]
  \begin{axis}[
      width=1.0\linewidth, 
      height=0.6\linewidth,
      grid=major,
      grid style={dashed,gray!30},
      xlabel=Layer,
      ylabel=$E(\bff^{(l)}) / E(\bff^{(0)})$,
      ylabel near ticks,
      legend style={at={(1.125,1.3)},anchor=north,scale=1.0, cells={anchor=west}, font=\large,},
      legend columns=-1,
      xtick={0,3,7,15,31,64},
      xticklabels = {1,4,8,16,32,64},
      yticklabel style={
        /pgf/number format/fixed,
        /pgf/number format/precision=3
      },
      scaled y ticks=false,
      every axis plot post/.style={ultra thick},
    ]
    \addplot[rgb color={255,0,0}]
    table[x=layer,y=coraGCN_omega1,col sep=comma] {data/EnergyValues.csv};
    \addplot[rgb color={175,0,0}, style=dashed]
    table[x=layer,y=coraGCN_omega01
    ,col sep=comma] {data/EnergyValues.csv};

    \addplot[rgb color={120,0,0}, style=dotted]
    table[x=layer,y=coraGCN_omega001
    ,col sep=comma] {data/EnergyValues.csv}; 
    
    \addplot[rgb color={50,0,0}, style=dashdotted]
    table[x=layer,y=coraGCN_omega0001
    ,col sep=comma] {data/EnergyValues.csv};

    \addplot[blue]
    table[x=layer,y=coraGCN_omegamix
    ,col sep=comma] {data/EnergyValues.csv}; 

    %\addplot[red, draw=none] coordinates {(1,1)};
    %\addplot[blue, draw=none] coordinates {(1,1)};
    %\addplot[gray, draw=none] coordinates {(1,1)};
    %\addplot[gray, style=dotted, draw=none] coordinates {(1,1)};
    \legend{$\omega=1$, $\omega=0.1$, $\omega=0.001$, $\omega=0.0001$, $\omega_{learnt}$}
    \end{axis}
\end{tikzpicture}
%\subcaption{$\omega$GCN.}
{$\omega$GCN}
\label{fig:energyOmegaGGCN_cora}
\end{minipage} \hspace{0.5em}
\begin{minipage}{.48\textwidth}
\centering
\begin{tikzpicture}[define rgb/.code={\definecolor{mycolor}{RGB}{#1}},
                    rgb color/.style={define rgb={#1},mycolor}]
  \begin{axis}[
      width=1.0\linewidth, 
      height=0.6\linewidth,
      grid=major,
      grid style={dashed,gray!30},
      xlabel=Layer,
      %ylabel=$E(\bff^{(l)}) / E(\bff^{(0)})$,
      %ylabel near ticks,
      legend style={at={(1.3,1.3)},anchor=north,scale=1.0, draw=none, cells={anchor=west}, font=\tiny, fill=none},
      legend columns=-1,
      xtick={0,3,7,15,31,64},
      xticklabels = {1,4,8,16,32,64},
      yticklabel style={
        /pgf/number format/fixed,
        /pgf/number format/precision=3
      },
      scaled y ticks=false,
      every axis plot post/.style={ultra thick},
    ]
    \addplot[rgb color={255,0,0}]
    table[x=layer,y=coraGAT_omega1,col sep=comma, forget plot] {data/EnergyValues.csv};
    
    \addplot[rgb color={175,0,0}, style=dashed]
    table[x=layer,y=coraGAT_omega01
    ,col sep=comma, forget plot] {data/EnergyValues.csv};

    \addplot[rgb color={120,0,0}, style=dotted]
    table[x=layer,y=coraGAT_omega001
    ,col sep=comma, forget plot] {data/EnergyValues.csv}; 
    
    \addplot[rgb color={50,0,0}, style=dashdotted]
    table[x=layer,y=coraGAT_omega0001
    ,col sep=comma, forget plot] {data/EnergyValues.csv};

    \addplot[blue]
    table[x=layer,y=coraGAT_omegamix
    ,col sep=comma, forget plot] {data/EnergyValues.csv}; 

    %\addplot[red, draw=none] coordinates {(1,1)};
    %\addplot[blue, draw=none] coordinates {(1,1)};
    %\addplot[gray, draw=none] coordinates {(1,1)};
    %\addplot[gray, style=dotted, draw=none] coordinates {(1,1)};
    \addlegendimage{empty legend}
    \addlegendimage{empty legend}
    \addlegendimage{empty legend}\addlegendentry{}
    %\legend{$\omega=1$, $\omega=0.1$, $\omega=0.001$, $\omega=0.0001$, $\omega_{learnt}$}
    \end{axis}
\end{tikzpicture}
%\subcaption{$\omega$GAT.}
{$\omega$GAT}
\label{fig:energyOmegaGAT_cora}
\end{minipage}
\caption{Node features energy at the $l$-th layer relative to the initial node embedding energy on Cora. Both $\omega$GCN and $\omega$GAT control the respective energies from \eqref{eq:dirichletEnergy} and \eqref{eq:gatEnergy} to avoid over-smoothing, while the baselines with $\omega=1$ reduce the energies to 0 and over-smooth. 
}
\label{fig:energyFig1}

\end{figure*}

%% file: images/FigEnergyCiteSeer_PubMed.tex
\begin{figure*}[h]
\centering
\begin{minipage}{0.48\textwidth}
\centering
\begin{tikzpicture}[define rgb/.code={\definecolor{mycolor}{RGB}{#1}},
                    rgb color/.style={define rgb={#1},mycolor}]
  \begin{axis}[
      width=1.0\linewidth, 
      height=0.6\linewidth,
      grid=major,
      grid style={dashed,gray!30},
      xlabel=Layer,
      ylabel=$E(\bff^{(l)}) / E(\bff^{(0)})$,
      ylabel near ticks,
      legend style={at={(1.125,1.3)},anchor=north,scale=1.0, cells={anchor=west}, font=\large,},
      legend columns=-1,
      xtick={0,3,7,15,31,64},
      xticklabels = {1,4,8,16,32,64},
      yticklabel style={
        /pgf/number format/fixed,
        /pgf/number format/precision=3
      },
      scaled y ticks=false,
      every axis plot post/.style={ultra thick},
    ]
    \addplot[rgb color={255,0,0}]
    table[x=layer,y=coraGAT_omega1,col sep=comma] {data/EnergyValues_Citeseer.csv};
    \addplot[rgb color={175,0,0}, style=dashed]
    table[x=layer,y=coraGAT_omega01
    ,col sep=comma] {data/EnergyValues_Citeseer.csv};

    \addplot[rgb color={120,0,0}, style=dotted]
    table[x=layer,y=coraGAT_omega001
    ,col sep=comma] {data/EnergyValues_Citeseer.csv}; 
    
    \addplot[rgb color={50,0,0}, style=dashdotted]
    table[x=layer,y=coraGAT_omega0001
    ,col sep=comma] {data/EnergyValues_Citeseer.csv};

    \addplot[blue]
    table[x=layer,y=coraGAT_omegamix
    ,col sep=comma] {data/EnergyValues_Citeseer.csv}; 

    %\addplot[red, draw=none] coordinates {(1,1)};
    %\addplot[blue, draw=none] coordinates {(1,1)};
    %\addplot[gray, draw=none] coordinates {(1,1)};
    %\addplot[gray, style=dotted, draw=none] coordinates {(1,1)};
    \legend{$\omega=1$, $\omega=0.1$, $\omega=0.001$, $\omega=0.0001$, $\omega_{learnt}$}
    \end{axis}
\end{tikzpicture}
{Citeseer}
\label{fig:energyOmegaGAT_citeseer}
\end{minipage} \hspace{0.5em}
\begin{minipage}{.48\textwidth}
\centering
\begin{tikzpicture}[define rgb/.code={\definecolor{mycolor}{RGB}{#1}},
                    rgb color/.style={define rgb={#1},mycolor}]
  \begin{axis}[
      width=1.0\linewidth, 
      height=0.6\linewidth,
      grid=major,
      grid style={dashed,gray!30},
      xlabel=Layer,
      %ylabel=$E(\bff^{(l)}) / E(\bff^{(0)})$,
      %ylabel near ticks,
      legend style={at={(1.3,1.3)},anchor=north,scale=1.0, draw=none, cells={anchor=west}, font=\tiny, fill=none},
      legend columns=-1,
      xtick={0,3,7,15,31,64},
      xticklabels = {1,4,8,16,32,64},
      yticklabel style={
        /pgf/number format/fixed,
        /pgf/number format/precision=3
      },
      scaled y ticks=false,
      every axis plot post/.style={ultra thick},
    ]
    \addplot[rgb color={255,0,0}]
    table[x=layer,y=coraGAT_omega1,col sep=comma, forget plot] {data/EnergyValues_Pubmed.csv};
    
    \addplot[rgb color={175,0,0}, style=dashed]
    table[x=layer,y=coraGAT_omega01
    ,col sep=comma, forget plot] {data/EnergyValues_Pubmed.csv};

    \addplot[rgb color={120,0,0}, style=dotted]
    table[x=layer,y=coraGAT_omega001
    ,col sep=comma, forget plot] {data/EnergyValues_Pubmed.csv}; 
    
    \addplot[rgb color={50,0,0}, style=dashdotted]
    table[x=layer,y=coraGAT_omega0001
    ,col sep=comma, forget plot] {data/EnergyValues_Pubmed.csv};

    \addplot[blue]
    table[x=layer,y=coraGAT_omegamix
    ,col sep=comma, forget plot] {data/EnergyValues_Pubmed.csv}; 

    %\addplot[red, draw=none] coordinates {(1,1)};
    %\addplot[blue, draw=none] coordinates {(1,1)};
    %\addplot[gray, draw=none] coordinates {(1,1)};
    %\addplot[gray, style=dotted, draw=none] coordinates {(1,1)};
    \addlegendimage{empty legend}
    \addlegendimage{empty legend}
    \addlegendimage{empty legend}\addlegendentry{}
    %\legend{$\omega=1$, $\omega=0.1$, $\omega=0.001$, $\omega=0.0001$, $\omega_{learnt}$}
    \end{axis}
\end{tikzpicture}
{Pubmed}
\label{fig:energyOmegaGAT_pubmed}
\end{minipage}
\caption{Node features energy at the $l$-th layer relative to the initial node embedding energy on Citeseer and Pubmed. $\omega$GAT controls the energy from Eq. \eqref{eq:gatEnergy} to avoid over-smoothing, while the baseline GAT with $\omega=1$ reduce the energy to 0 and over-smooth. 
}
\label{fig:appendixGATOVERSMOOTHING}

\end{figure*}